\documentclass[10pt,twocolumn,letterpaper]{article}

\usepackage[pagenumbers]{wacv} 

\usepackage{graphicx}
\usepackage{amsmath,amssymb}
\usepackage{booktabs}
\usepackage{pifont}
\usepackage{multirow,multicol}
\usepackage{array}
\usepackage{textpos}
\usepackage[pagebackref,breaklinks,colorlinks]{hyperref}
\usepackage[edges]{forest}

\makeatletter
\newcommand{\copyrightnotice}[1]{\gdef\@copyrightnotice{#1}}
\newcommand{\toappear}[1]{\gdef\@toappear{#1}}
\let\@copyrightnotice\@empty
\let\@toappear\@empty
\newcommand{\ps@myfooter}{%
  \let\@mkboth\@gobbletwo
  \def\@oddhead{}%
  \def\@evenhead{}%
  \def\@oddfoot{\rlap{\@toappear}\hfil\thepage\hfil\llap{\@copyrightnotice}}%
  \let\@evenfoot\@oddfoot
}
\makeatother

\toappear{\it To appear in Proc.\ WACV2025}
\copyrightnotice{\copyright\ 2025 IEEE}
\usepackage{ifthen}
\usepackage{atbegshi}
\newcounter{pagecounter}
\AtBeginShipout{%
  \stepcounter{pagecounter}%
  \ifthenelse{\value{pagecounter}=1}{%
    \AtBeginShipoutUpperLeft{%
      \put(\dimexpr\paperwidth-20.3cm\relax,-0.8cm){%
        \parbox[t]{19cm}{%
          \copyright\ 2025 IEEE. Personal use of this material is permitted. 
          Permission from IEEE must be obtained for all other uses, in any current or future media, including reprinting/republishing this material for advertising or promotional purposes, creating new collective works, for resale or redistribution to servers or lists, or reuse of any copyrighted component of this work in other works.%
        }%
      }%
    }%
  }{}%
}

\definecolor{foldercolor}{RGB}{124,166,198}
\definecolor{folderbg}{RGB}{124,166,198}
\definecolor{folderborder}{RGB}{110,144,169}

\def\Size{4pt}
\tikzset{folder/.pic={\filldraw[draw=folderborder,top color=folderbg!50,bottom color=folderbg]
(-1.05*\Size,0.2\Size+5pt) rectangle ++(.75*\Size,-0.2\Size-5pt);  
\filldraw[draw=folderborder,top color=folderbg!50,bottom color=folderbg]
(-1.15*\Size,-\Size) rectangle (1.15*\Size,\Size);
  }
}

\forestset{is file/.style={edge path'/.expanded={%
        ([xshift=\forestregister{folder indent}]!u.parent anchor) |- (.child anchor)},
        inner sep=1pt},
    this folder size/.style={edge path'/.expanded={%
        ([xshift=\forestregister{folder indent}]!u.parent anchor) |- (.child anchor) pic[solid]{folder=#1}}, inner ysep=0.2*#1},
    folder tree indent/.style={before computing xy={l=#1}},
    folder icons/.style={folder, this folder size=#1, folder tree indent=3*#1},
    folder icons/.default={10pt},
}

\usepackage[capitalize]{cleveref}
\crefname{section}{Sec.}{Secs.}
\Crefname{section}{Section}{Sections}
\Crefname{table}{Table}{Tables}
\crefname{table}{Tab.}{Tabs.}

\def\wacvPaperID{xxx} 
\def\confName{WACV}
\def\confYear{2025}

\begin{document}

\title{SynDroneVision: A Synthetic Dataset for Image-Based Drone Detection}

\author{
    Tamara R. Lenhard$^{1,2}$ 
    \and
    Andreas Weinmann$^{2}$
    \and
    Kai Franke$^{1}$
    \and
    Tobias Koch$^{1}$\\\vspace{-0.2cm}\and
    $^{1}$Institute for the Protection of Terrestrial Infrastructures,
    German Aerospace Center (DLR)\\\vspace{-0.6cm}\and
    $^{2}$Working Group Algorithms for Computer Vision, Imaging and Data Analysis,\\ University of Applied Sciences Darmstadt\\\and
    {\tt\small \{tamara.lenhard, kai.franke, tobias.koch\}@dlr.de, andreas.weinmann@h-da.de}
}

\maketitle

\begin{abstract}
Developing robust drone detection systems is often constrained by the limited availability of large-scale annotated training data and the high costs associated with real-world data collection. However, leveraging synthetic data generated via game engine-based simulations provides a promising and cost-effective solution to overcome this issue. Therefore, we present SynDroneVision, a synthetic dataset specifically designed for RGB-based drone detection in surveillance applications. Featuring diverse backgrounds, lighting conditions, and drone models, SynDroneVision offers a comprehensive training foundation for deep learning algorithms. To evaluate the dataset's effectiveness, we perform a comparative analysis across a selection of recent YOLO detection models. Our findings demonstrate that SynDroneVision is a valuable resource for real-world data enrichment, achieving notable enhancements in model performance and robustness, while significantly reducing the time and costs of real-world data acquisition. SynDroneVision will be publicly released upon paper acceptance.
\end{abstract}

\section{Introduction}
\label{sec:intro}
Unmanned aerial vehicles (UAVs), commonly known as drones, have become integral to a variety of sectors, including agriculture, logistics, surveillance, and recreation. However, their rapid proliferation introduces new challenges, particularly in terms of security and privacy protection~\cite{Dafrallah:2024}. Therefore, the implementation of effective drone detection systems is crucial to mitigate the risks associated with unauthorized or malicious drone activities. Combining optical sensors, specifically cameras, with advanced deep learning (DL) techniques represents a highly promising and economically efficient detection strategy~\cite{Elsayed:2021}. Nevertheless, the effectiveness of DL models is heavily reliant on extensive and diverse training data~\cite{Singh:2024,Yu:2023}.

In practical applications, the acquisition of substantial amounts of annotated real-world data is both time-consuming and resource-intensive~\cite{Jiang:2023,Svanstroem:2021}. Additional constraints, such as non-fly zones and adverse weather conditions, further complicate the data collection process. Leveraging synthetic data presents a viable alternative to circumvent environmental limitations and significantly reduce acquisition costs~\cite{Dieter:2023_12,Singh:2024} (not only in drone detection, but also in other domains~\cite{Aboli:2023,Kloukiniotis:2022,Anagnostopoulou:2023,Tran:2024}). In particular, the application of game engine-based data generation techniques enables the efficient and physically precise simulation of diverse real-world conditions~\cite{Marez:2020,Dieter:2023_12,Kolbeinsson:2024,Black:2023}. It facilitates the seamless interchange of environmental configurations (e.g., from urban landscapes to rural terrains), offering the potential for comprehensive coverage of diverse scenarios, including those inadequately represented by real-world data. Furthermore, the ability to rapidly alter illumination, time of day, and weather conditions -- from clear summer days to overcast skies within seconds~-- provides a time-efficient solution without compromising data diversity. A broad selection of interchangeable drone models, materials, and textures supports a high variability in drone appearances, in contrast to the often limited drone selection in practical applications~\cite{Aksoy:2019,Zheng:2021}. Moreover, a key advantage of synthetic data is the automated generation of pixel-precise annotations~\cite{Lenhard:2024}. This capability accelerates both training and validation processes, enabling more rapid experimentation and iteration cycles. Furthermore, it significantly decreases resource requirements associated with traditional data collection methods~\cite{Jiang:2023}, particularly in terms of annotation costs and recording time.

Despite the substantial benefits of synthetic data, there is still a gap between simulated scenarios and real-world conditions~\cite{Barisic:2022,Marez:2020}. This discrepancy can negatively impact detection quality, especially when transferring drone detection models trained exclusively on synthetic data to real-world applications. However, leveraging hybrid datasets -- incorporating both real and synthetic data, with real data shares $<1\%$ -- has proven to be an effective training strategy to overcome this issue~\cite{Symeonidis:2022,Dieter:2023}. 

Although the generation of synthetic data is a topic of ongoing research, particularly in the field of drone detection, there is currently only one publicly available synthetic dataset: the S-UAV-T dataset by Barisic \etal~\cite{Barisic:2022}. This dataset is specifically designed for UAV-to-UAV detection, featuring perspectives that deviate from standard surveillance configurations. 

\textbf{Contributions.} Addressing the scarcity of publicly available synthetic datasets for image-based drone detection, we introduce \textit{SynDroneVision} -- a comprehensive dataset featuring diverse environments, drone models, and lighting conditions. We provide a comparative analysis of state-of-the-art models, demonstrating the effectiveness of SynDroneVision, especially in combination with real-world data. Furthermore, we assess the robustness of this approach using out-of-distribution data.

The remainder of this paper is organized as follows: \Cref{sec:relatedwork} provides an overview of related research on image-based drone detection and publicly available datasets. \Cref{sec:syndronevision} details the generation process, defining features, and the composition of SynDroneVision. The experimental design and results are outlined in \Cref{sec:experiment}. Conclusions are presented in \Cref{sec:conclusion}.

\begin{table*}[t!]
\caption{Comprehensive information on publicly available datasets for image-based drone detection, encompassing both real and synthetic data. The symbols \ding{55} (does not apply) and \ding{51} (applies) indicate the presence of image sequences (4th column from the right), the inclusion of diverse drone models (3rd column from the right), and the incorporation of distractor objects (2nd column from the right).}
  \label{tab:dronedatasets}
  \centering
  \begin{tabular}{@{}|l|c|c|c|c|c|c|c|c|@{}}
    \hline
    Dataset & \multicolumn{4}{c|}{No. Images} & Img. & Diff. & Dist. & Max. Img.\\
    & {\small train} & {\small val} & {\small test} & {\small total} & Sequ. & Drones & Objs. & Resolution\\\hline
    
    USC Drone Detect. \& Track.~\cite{Chen:2017,Wang:2018}  & -- & -- & -- & \ \ \ 27,000$^{\text{\ding{72}}}$ & \ding{51} & \ding{55} & \ding{55} & 1920$\times$1080\\\hline
    
    Drone Dataset~\cite{Aksoy:2019} & 3,611 & -- & 401 & \ \ 4,012 & \ding{55} & \ding{55} & \ding{55} & 3840$\times$2160\\\hline
    
    MAV-VID~\cite{Rodriguez-Ramos:2020} & 29,500 & 10,732 & -- & 40,232 & \ding{51} & \ding{51} & \ding{55} & --\\\hline
    
    Det-Fly~\cite{Zheng:2021} & -- & -- & -- & \ \ 13,271$^{\text{\ding{72}}}$ & (\ding{51}) & \ding{55} & \ding{55} & 3840$\times$2160\\\hline
    
    UAV-Eagle~\cite{Barisic:2021} & -- & -- & -- & 510$^{\text{\ding{72}}}$ & \ding{51} & \ding{55} & \ding{55} & 1920$\times$1080\\\hline
    
    UAVData~\cite{Zeng:2021} & -- & -- & -- & \ \ 13,803$^{\text{\ding{72}}}$ & \ding{51} & \ding{51}  & \ding{51} & 1280$\times$720\\\hline
    
    Halmstadt Data~\cite{Svanstroem:2021} & -- & -- & -- & 203,328$^{\text{\ding{115}}}$ & \ding{51} & \ding{51} & \ding{51} & 640$\times$512\\\hline
    
    DUT Anti-UAV~\cite{Zhao:2022} & 5,200 & 2,600 & 2,200 & 10,000 & (\ding{51}) & \ding{51} & \ding{55} & 5616$\times$3744 \\ \hline
    
    VisioDECT~\cite{Ajakwe:2022} & -- & -- & -- & \ \ 20,924$^{\text{\ding{72}}}$ & \ding{51}  & \ding{51} & \ding{55} & 852$\times$480\\\hline
    
    Malicious Drones~\cite{Jamil:2022} & 543 & -- & 233 & 776 & \ding{55} & \ding{51} & \ding{51} & 224$\times$224\\\hline
    
    S-UAV-T~\cite{Barisic:2022} \textit{(synthetic)} & -- & -- & -- & \ \ 52,500$^{\text{\ding{72}}}$ & \ding{55} & \ding{51} & \ding{55} & 608$\times$608 \\ \hline
    
    Drone-vs-Bird Detection Ch.~\cite{Coluccia:2024} & 85,904 & -- & \ \ --$^{\text{\ding{91}}}$ & 85,904 & \ding{51} & \ding{51} & \ding{51} & 3840$\times$2160\\ \hline
    
    \multicolumn{9}{l}{
    {\footnotesize\ding{91}}{\small \ not publicly available}
    \hspace{0.3cm}
    {\footnotesize\ding{72}}{\small \ no subdivision into train, val, and test}
    \hspace{0.3cm}
    {\footnotesize\ding{115}}{\small \ RGB + IR data}
    }
  \end{tabular}
\end{table*}

\section{Related Work}
\label{sec:relatedwork}
In the following sections, we briefly summarize recent advances in image-based drone detection and assess the characteristics of publicly available drone detection datasets, emphasizing their similarities and differences.

\subsection{Drone Detection}
State-of-the-art techniques for RGB-based drone detection primarily leverage single-stage DL algorithms, which offer an optimal balance between real-time performance and precision. A majority of methodologies employ variants of the You Only Look Once (YOLO) models, including YOLOv3, YOLOv5, and YOLOv8, either in their original configurations~\cite{Barisic:2022,Munir:2023} or with custom modifications~\cite{Lv:2022,Lenhard:2024,Huang:2024}. Architectural innovations typically seek to resolve particular challenges encountered in drone detection, including the identification of small drones~\cite{Lv:2022,Liu:2021,Jiang:2023,Huang:2024}, the differentiation between drones and other aerial entities (e.g., birds)~\cite{Lv:2022,Coluccia:2024}, and the mitigation of camouflage effects~\cite{Lenhard:2024}. Transformer-based approaches offer an effective, albeit less frequently employed, alternative to traditional detection techniques~\cite{Jamil:2022}.

Training drone detection models predominantly relies on (self-collected)  application-specific real-world data. However, significant efforts are also directed towards the creation and utilization of synthetic data (e.g., see~\cite{Barisic:2022,Marez:2020}). Despite variations in generation techniques, prevailing research highlights the substantial potential of synthetic data, particularly in combination with real-world data. Prevalent training strategies for improving detection quality by integrating real and synthetic data include mixed-data training~\cite{Chen:2017,Symeonidis:2022} and fine-tuning models, initially trained on synthetic data, with real-world data~\cite{Barisic:2022,Marez:2020}. However, the optimal ratio of synthetic to real-world data is a controversial topic of ongoing research\cite{Dieter:2023}.

\subsection{Datasets}
\label{subsec:datasets}
Publicly available datasets for drone detection can be divided into two primary categories. The first category includes datasets exclusively designed for detection tasks~\cite{Barisic:2022,Barisic:2021,Aksoy:2019}, typically featuring individual images. The second category comprises datasets that support both detection and tracking~\cite{Chen:2017,Wang:2018,Zhao:2022,Coluccia:2024}, generally including sequential data or specialized subsets featuring both image and video files. Except for the dataset by Barisic \etal~\cite{Barisic:2022}, the majority of datasets consists of real-world data sourced from Google images~\cite{Aksoy:2019}, YouTube videos~\cite{Aksoy:2019}, or self-recorded footage~\cite{Chen:2017,Wang:2018,Zeng:2021} using static, moving, handheld, or drone-mounted devices. Beyond the variability in data origins and collection techniques, available datasets exhibit further variations in the following attributes:

\begin{itemize}
    \item \textit{Dataset Size} -- The costly nature of real-world data acquisition leads to significant discrepancies in the sizes of publicly available datasets. For instance, the datasets UAV-Eagle~\cite{Barisic:2021} and Malicious Drones~\cite{Jamil:2022} are relatively small, with 510 and 776 images, respectively (see \Cref{tab:dronedatasets}). Most datasets comprise 4,000~\cite{Aksoy:2019} to 40,232 images~\cite{Rodriguez-Ramos:2020}. Exceptions include the Halmstad Data~\cite{Svanstroem:2021}, featuring 203,328 annotated frames (IR + RGB), and the Drone-vs-Bird Detection Challenge dataset~\cite{Coluccia:2024}, with 85,904 annotated frames. (Notably, the Drone-vs-Bird Detection Challenge dataset~\cite{Coluccia:2024} constitutes a comprehensive compilation of data, collected over time and continually enhanced by input from various contributors.)

    \item \textit{Image Resolution} -- Publicly available drone detection datasets encompass a wide spectrum of resolutions, ranging from low-resolution (e.g., 224$\times$224 pixels~\cite{Jamil:2022} and 608$\times$608 pixels~\cite{Barisic:2022}) to high-resolution images (e.g., 5616$\times$3744  pixels~\cite{Zhao:2022}). The variability in resolution is observed both across different datasets and within individual datasets. While some datasets maintain a uniform resolution~\cite{Jamil:2022,Svanstroem:2021,Barisic:2022,Zheng:2021,Chen:2017,Wang:2018,Ajakwe:2022,Barisic:2021}, others offer a diverse range of resolutions~\cite{Zhao:2022,Coluccia:2024,Aksoy:2019}. For instance, the DUT Anti-UAV dataset provided by Zhao \etal~\cite{Zhao:2022} includes images with resolutions varying from 240$\times$160 to 5616$\times$3744 pixels, whereas the Det-Fly dataset by 
    Zheng \etal~\cite{Zheng:2021} is characterized by a consistent resolution of 3840$\times$2160 pixels.
    
    \item \textit{Drone Models, Size \& Position} -- The representation of drone models across various datasets exhibits considerable heterogeneity. While some datasets are restricted to a single drone model~\cite{Rodriguez-Ramos:2020,Zheng:2021,Aksoy:2019}, others encompass multiple models~\cite{Coluccia:2024,Zeng:2021,Ajakwe:2022,Svanstroem:2021,Zhao:2022}, typically comprising three~\cite{Svanstroem:2021} to eight~\cite{Coluccia:2024} distinct types. Exceptions include the DUT Anti-UAV dataset~\cite{Zhao:2022}, featuring 35 drone models, and the synthetic dataset by Barisic  \etal~\cite{Barisic:2022}. Additionally, variability is observed in the number of drones per image (ranging from single~\cite{Rodriguez-Ramos:2020,Barisic:2021} to multiple drones~\cite{Zeng:2021,Zhao:2022,Barisic:2022} per frame), as well as in their size and position. Nevertheless, a common feature across most datasets is the prevalence of small, centrally positioned drones, typically in conventional colors such as black and~/~or white.
    
    \item \textit{Distractor Objects} -- In addition to drones, some datasets encompass other (drone-like) objects~\cite{Aksoy:2019}, including birds~\cite{Svanstroem:2021,Coluccia:2024,Jamil:2022}, airplanes~\cite{Svanstroem:2021,Jamil:2022}, helicopters~\cite{Svanstroem:2021,Jamil:2022}, and balloons~\cite{Zeng:2021}. These distractor objects are either explicitly annotated~\cite{Svanstroem:2021,Jamil:2022} or incorporated in a more implicit manner~\cite{Coluccia:2024,Zeng:2021,Aksoy:2019}.

    \item \textit{Backgrounds} -- The majority of datasets feature outdoor environments such as urban landscapes, forests, farmland, airports, and coastal areas across different regions around the globe (e.g., Sweden~\cite{Svanstroem:2021} vs. China~\cite{Zhao:2022}). The visual compositions comprise an assortment of elements, including skies, buildings, playgrounds, vegetation, and other landscape features, captured from diverse viewing angles (e.g., top-down and bottom-up, as in~\cite{Zhao:2022}). An exception is the dataset by Zeng \etal~\cite{Zeng:2021} which also features indoor scenes. 
    
    \item \textit{Illumination \& Weather} -- Real-world datasets are predominantly recorded in daylight~\cite{Svanstroem:2021,Zhao:2022,Zeng:2021} and feature diverse weather conditions~\cite{Coluccia:2024, Jamil:2022} (including cloudy~\cite{Ajakwe:2022, Zhao:2022}, sunny~\cite{Ajakwe:2022, Zhao:2022}, and snowy~\cite{Zhao:2022}). Some datasets also account for low-light conditions such as night, dawn, and dusk~\cite{Ajakwe:2022,Zhao:2022,Coluccia:2024}. Reflecting the unpredictable nature of real-world lighting, these datasets often (unintentionally) exhibit rapid illumination changes or direct sun glare~\cite{Zeng:2021, Coluccia:2024}. In contrast, the synthetic dataset by Barisic \etal~\cite{Barisic:2022} is characterized by more controlled lighting conditions, including daylight and twilight.
    
    \item \textit{Annotation Process} -- Annotations are typically generated manually by individual experts~\cite{Rodriguez-Ramos:2020} or teams~\cite{Ajakwe:2022}. An exception is the synthetic dataset by Barisic \etal~\cite{Barisic:2022}, which employs an automated annotation pipeline based on Blender~\cite{Blender} and Cycles~\cite{Cycles}. Manual annotation techniques often suffer from inconsistencies in terms of quality and precision. Additionally, variations in file formats (e.g., .txt~\cite{Barisic:2021} vs.~.mat~\cite{Svanstroem:2021}) and bounding box definitions further compromise the datasets' practical applicability. Methodologies, tools, and quality control measures are often insufficiently documented, resulting in a lack of transparency.
\end{itemize}

In addition to RGB data, some datasets include other imaging modalities, such as infrared (IR) data~\cite{Svanstroem:2021}. \Cref{tab:dronedatasets} provides an overview of existing datasets and their characteristics, with further details in the supplementary material.

\section{SynDroneVision Dataset}
\label{sec:syndronevision}
This section provides a comprehensive overview of the proposed SynDroneVision dataset, detailing the employed generation technique, simulation parameter variations, its composition, and inherent characteristic properties.

\subsection{Data Generation Process}
\label{subsec:datagen}
The synthetic RGB data of the proposed dataset is generated using an advanced iteration of the data generation pipeline introduced by Dieter \etal~\cite{Dieter:2023_12}. Unlike Dieter \etal's pipeline –- reliant on Microsoft AirSim~\cite{Airsim} and Unreal Engine 4.25~\cite{Unreal} –- the employed pipeline leverages Colosseum~\cite{Colosseum} (the successor to Microsoft AirSim) and Unreal Engine 5.0. Unreal Engine 5.0 introduces advanced capabilities for rendering dynamic global illumination and reflections through the implementation of the fully dynamic global illumination and reflections system Lumen~\cite{Lumen}. This advancement significantly enhances the realism of depicted scenes (especially in terms of daytime-dependent light and shadow variations), thereby elevating the fidelity of synthetic RGB data. To ensure precise representation of Lumen-based lighting effects and reflections, we refine the data generation process by modifying the pipeline's data generation module (cf.~\cite{Dieter:2023_12}). In Dieter \etal's pipeline, visual sensor data was acquired through Scene Capture 2D Actors. However, when integrated with Unreal Engine 5.0, these components lack the ability to accurately capture the intricate details of Lumen's dynamic global illumination and reflections. To address this limitation, we implement the capture of RGB data via high resolution screenshots. All other pipeline components remain consistent with~\cite{Dieter:2023_12}. For detailed information on individual components, refer to~\cite{Dieter:2023_12}.

\begin{figure}[t!]
  \centering
  \includegraphics[width=0.48\textwidth, trim={0.1cm 8.9cm 0.1cm 0cm}, clip]{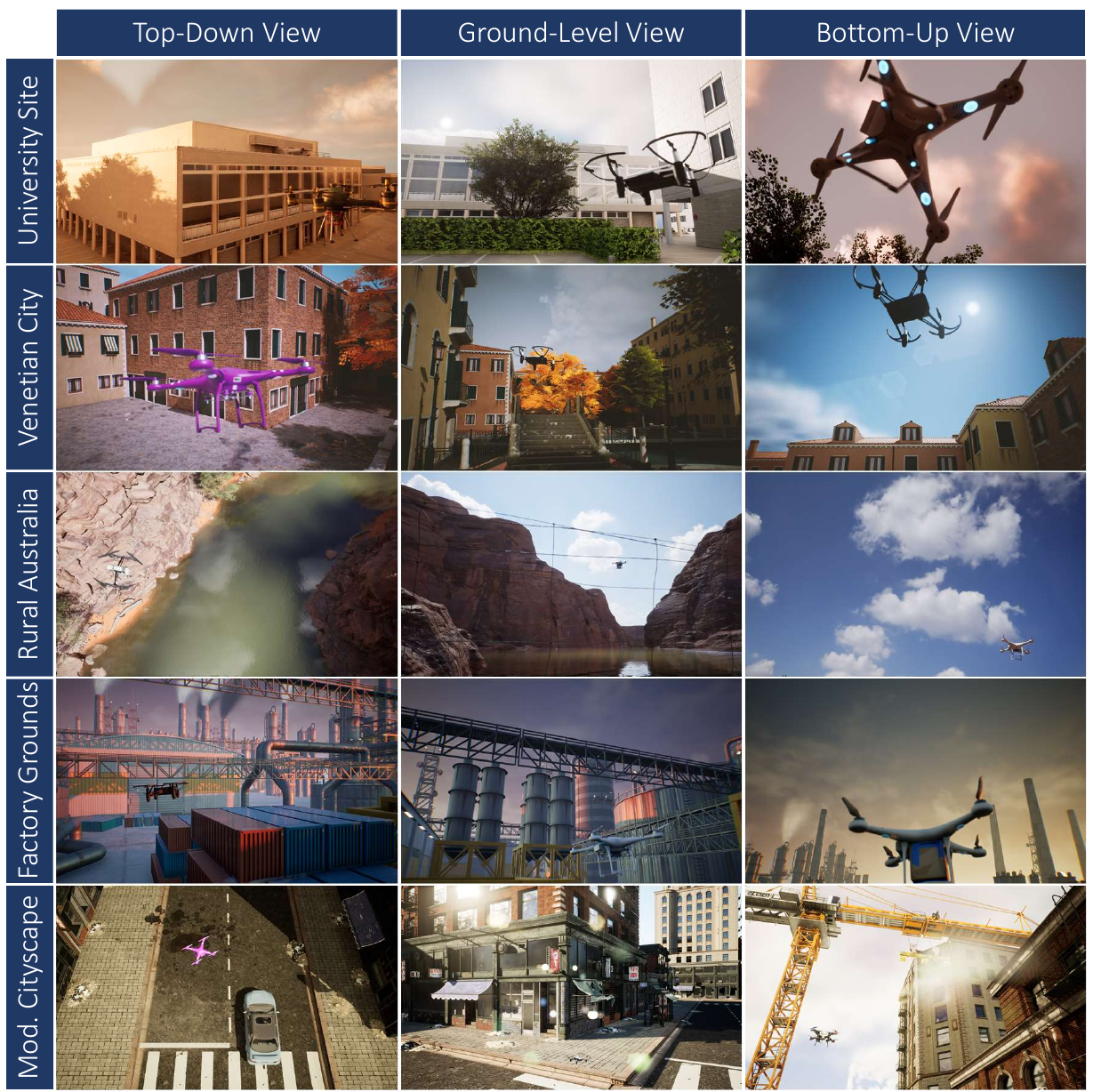}
  \caption{\label{fig:FOVs} A selection of synthetic images captured from diverse virtual environments  -- University Site (row 1), Venetian City (row 2), Rural Australia (row 3), Factory Grounds (row 4), and Modular Cityscape (last row) -- demonstrating SynDroneVision's diversity in terms of environmental conditions and camera perspectives (top-down, ground-level, and bottom-up).}
\end{figure}

The data acquisition process itself is predicated on a strategic placement of stationary virtual camera sensors, whose position and orientation are pre-determined with respect to the underlying simulation environment (inspired by a typical surveillance setup, cf. \Cref{fig:FOVs}). Data collection is systematically performed from each designated camera in a sequential manner, adhering to a pre-defined recording duration. The recording duration is synchronized with the drone's flight time. The drone's flight trajectory is determined by a probabilistic selection of waypoints, randomly positioned within the camera's field of view (up to 30 meters from its vantage point).

\subsection{Simulation Parameter and Domain Variations}
To foster a high level of diversity in the generated data, we introduce variations across multiple simulation components, including the environment, drone models, and lighting conditions. Detailed information on these aspects is provided in the following sections.

\begin{figure}[t!]
  \centering
  \includegraphics[width=0.48\textwidth]{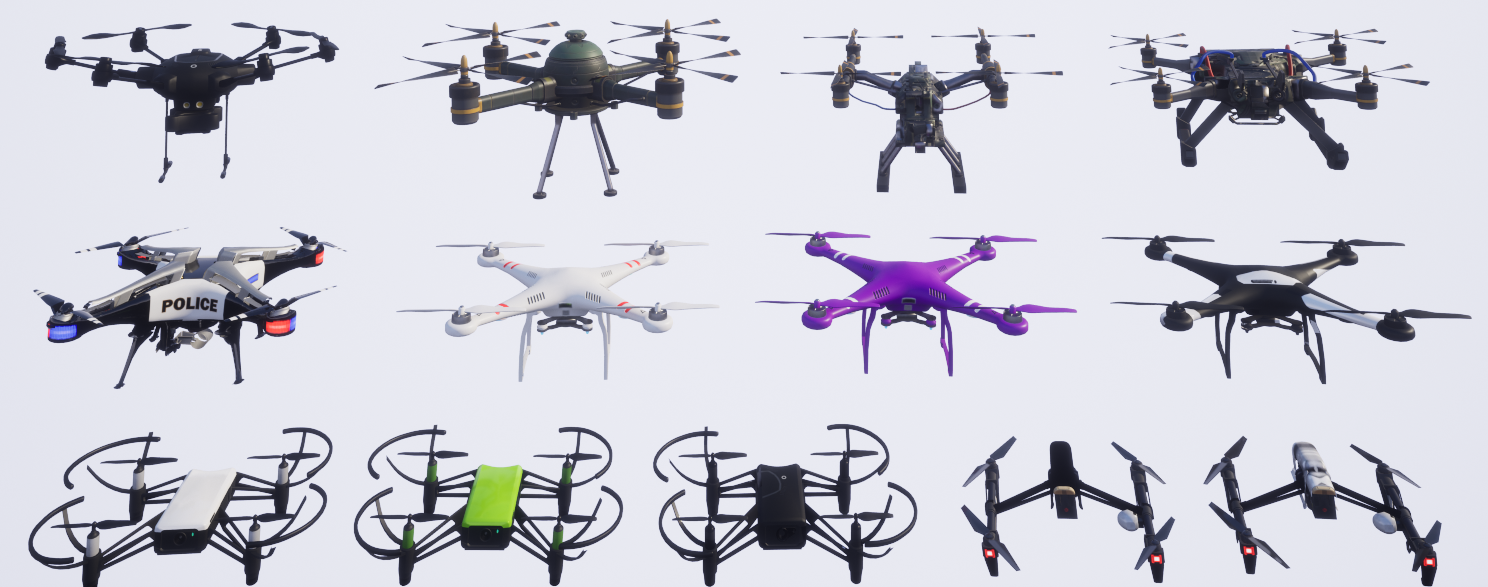}
  \caption{\label{fig:dronemodels} Drone models from~\cite{Unreal:QuadcopterPack,Unreal:MilitaryDrones,Unreal:DronePack} employed in the generation of the SynDroneVision dataset.}
\end{figure}

\subsubsection{Environments} To establish a foundation for simulating physically-realistic drone flights, we leverage a variety of three-dimensional environments, encompassing both commercially licensed and freely available options. The selection of environments is guided by the defining attributes of the real-world settings utilized in creating the DUT Anti-UAV dataset~\cite{Zhao:2022}. Particular emphasis is placed on incorporating environments with diverse complexity levels and substantial variations (cf. \Cref{fig:FOVs}), ensuring thorough data diversification. The employed environments and settings include: University Site, Venetian City~\cite{Unreal:Venice}, Farming Grounds~\cite{Unreal:Farming}, Rural Australia~\cite{Unreal:Australia}, City Park~\cite{Unreal:CityPark}, Factory Grounds~\cite{Unreal:Factory}, Urban Downtown~\cite{Unreal:DowntownWest}, and Modular Cityscape~\cite{Unreal:ModBuilding}. In publicly accessible environments (commercial or free), we utilized pre-existing demo maps (with minor modifications, cf. \Cref{subsubsec:illumination}) for data generation. Note that the environments predominantly consist of static geometry, with two notable exceptions: the shader-based animation of foliage and the dynamic motion of the drone. A detailed overview of the environments and their inherent characteristics can be found in the supplementary material (\Cref{tab:envs} and \Cref{fig:FOVs2}).

\subsubsection{Drone Models} To ensure a high degree of diversity in drone representation, we employ a selection of drone models from the commercially available Quadcopter Pack~\cite{Unreal:QuadcopterPack}, Drone Pack~\cite{Unreal:DronePack}, and Military Drone Pack~\cite{Unreal:MilitaryDrones} for synthetic data generation (see \Cref{fig:dronemodels}). Our selection features a variety of widely-deployed drone models, including the DJI Phantom (\Cref{fig:dronemodels}, second row, second model from the left) and the DJI Tello Ryze (\Cref{fig:dronemodels}, last row, third model from the left). Each drone model is rendered with realistic textures sourced from the respective asset packages. This contrasts with the approach of Barisic \etal~\cite{Barisic:2022}, characterized by texture randomization featuring unconventional drone textures.

\subsubsection{Illumination}\label{subsubsec:illumination} To incorporate a variety of realistic illumination conditions, we utilize the dynamic illumination and reflection system, Lumen~\cite{Lumen}, in conjunction with the Sun and Sky Actor~\cite{SunSkyActor} and the Post Process Volume~\cite{PostProcess} provided by Unreal Engine. The Sun and Sky Actor offers precise control over the sun's positioning based on geographic location, date, and time, while the Post Process Volume provides a comprehensive toolkit for regulating visual aesthetics and atmospheric properties (e.g., color grading, contrast, or bloom). By leveraging the dynamic nature of directional lights within the Sun and Sky Actor, lightmap baking becomes obsolete, thus eliminating the need for pre-computed lighting. Consequently, this configuration enables the creation of authentic renderings, accurately portraying the interplay of sunlight and shadow. 

During the data generation process, we systematically introduce variations in the intensity of the Directional Light Actor and the Rayleigh scattering properties of the Sky Atmosphere -- both fundamental components of the Sun and Sky Actor. Additionally, adjustments are made to the color temperature parameter within the Post Process Volume to further refine atmospheric properties. Thus, SynDroneVision offers a broad spectrum of illumination conditions, ranging from dawn (\Cref{fig:FOVs}, first row, first image) to dusk (\Cref{fig:FOVs}, first row, third image), from clear blue skies  (\Cref{fig:FOVs}, second row, third image) to overcast conditions (\Cref{fig:FOVs}, first row, second image). Note that the combination of Lumen, Sun and Sky Actor, and Post Process Volume is employed exclusively for data generation within the following environments: University Site, Venetian City~\cite{Unreal:Venice}, Farming Grounds~\cite{Unreal:Farming}, and Modular Cityscape~\cite{Unreal:ModBuilding}. The default illumination setup included in the environments Rural Australia~\cite{Unreal:Australia}, City Park~\cite{Unreal:CityPark}, Factory Grounds~\cite{Unreal:Factory}, and Urban Downtown~\cite{Unreal:DowntownWest} remains unchanged, as it already features a sophisticated (Lumen-based) implementation of lighting and reflections. Further details on illumination parameters are provided in the supplementary material. 

\begin{table*}[ht!]
\centering
\caption{Comprehensive composition overview of the SynDroneVision dataset.}
  \label{tab:synthdata}
  \begin{tabular}{@{}|l|c|c|c|c|c|c|@{}}
    \hline
    Environment & \multicolumn{4}{c|}{Image Count} & Camera & Drone\\
    & {\small train} & {\small val} & {\small test} & {\small total} & Positions &  Models\\
    \hline
    University Site & 29,111 & 1,000 & 500 & 30,611 & 13 & 8 \\\hline
    Venetian City & 19,566 & 1,000 & 500 & 21,066 & 8 & 6\\\hline
    Farming Grounds & 9,053 & 1,000 & 500 & 10,553 & 6 & 5 \\\hline
    Rural Australia & 11,589 & 1,000 & 500 & 13,089 & 6 & 5 \\\hline
    City Park & 7,310 & 1,000 & 500 &  8,810 & 4 & 4\\\hline
    Factory Grounds & 13,759 & 1,000 & 500 & 15,259 & 7 & 6 \\\hline
    Urban Downtown & 14,404 & 1,000 & 500 & 15,904 & 9 & 6\\\hline
    Modular Cityscape & 14,515 & 1,000 & 500 & 16,015 & 5 & 6\\\hline
    \textbf{Total} & \textbf{119,307}~/~\textbf{131,238}$^{\text{\ding{72}}}$  & \textbf{8,000}~/~\textbf{8,800}$^{\text{\ding{72}}}$  & \textbf{4,000} & \textbf{131,307}~/~\textbf{140,038}$^{\text{\ding{72}}}$  & \textbf{58} &  \textbf{13} \\\hline
     \multicolumn{3}{l}{
    {\footnotesize\ding{72}}{\small \ incl. blurring}}
    \end{tabular}
\end{table*}

\begin{table*}[t!]
\caption{Comprehensive details regarding the area and aspect ratios of objects featured in SynDroneVision -- across training, validation, and test splits -- in comparison to DUT Anti-UAV~\cite{Zhao:2022}.}
  \label{tab:characteristics}
  \centering
  \begin{tabular}{@{}|l|c|c|c|c|c|c|c|c|c|c|c|c|@{}}
    \cline{2-13}
    \multicolumn{1}{c|}{} &  \multicolumn{6}{c|}{SynDroneVision (Ours)} &  \multicolumn{6}{c|}{DUT Anti-UAV}\\\hline
    Split & \multicolumn{3}{c|}{Object Area Ratio} & \multicolumn{3}{c|}{Object Aspect Ratio} & \multicolumn{3}{c|}{Object Area Ratio} & \multicolumn{3}{c|}{Object Aspect Ratio}\\
    & {\small min} & {\small avg.}  & {\small max} & {\small min} & {\small avg.}  & {\small max} & {\small min} & {\small avg.}  & {\small max} & {\small min} & {\small avg.}  & {\small max}\\\hline
    train & 0.001 & 0.322 & 1.0 & 0.021 & 1.291 & 9.993 & 0.000026 & 0.013 & 0.700 & 1.0 & 1.910 & 5.420\\\hline
    val   & 0.006 & 0.323 & 1.0 & 0.020 & 1.330 & 9.855 & 0.000002 & 0.013 & 0.690 & 1.0 & 1.910 & 6.670\\\hline
    test  & 0.009 & 0.323 & 1.0 & 0.041 & 1.302 & 8.383 & 0.000041 & 0.014 & 0.470 & 1.0 & 1.920 & 5.090\\\hline
  \end{tabular}
    
\end{table*}

\subsection{Post-Processing} \label{subsubsec:blur}
To further increase the data diversity, a subset of randomly selected images undergoes post-capture blurring. Considering a synthetically generated image $\mathbf{I}\in  \mathbb{R}^{W\times H\times 3}$, where $W\in\mathbb{N}$ denotes the width and $H\in\mathbb{N}$ denotes the height, the blurring procedure is defined by the following convolution operation:\vspace*{-0.1cm}
\begin{equation}
  \mathbf{I}'(x,y) = \sum_{i=0}^{2m} \sum_{j=0}^{2n} \mathbf{I}(x+i-m, y+j-n) \cdot \mathbf{K}(i, j)
  \label{eq:blurring}
\end{equation}
\noindent where $x\in\{0,...,W-1\}$ and $y\in\{0,...,H-1\}$. The kernel $\mathbf{K}\in\mathbb{R}^{M\times N}$ is characterized by the dimensions $M=2m-1$ and $N=2n-1$ with $m,n\in\mathbb{N}$. In our application, the kernel is specified as either an average kernel or a Gaussian kernel. The average kernel is given by $\mathbf{K}(i, j)=\frac{1}{(2m-1)(2n-1)}$, with indices $i\in\{0,...,2m\}$ and $j\in\{0,...,2n\}$ determining the kernel's spatial extent. Conversely, the Gaussian kernel is  described by $\mathbf{K}(i, j)=\frac{1}{(2\pi \sigma^2)} \exp \left( -\frac{(i-m)^2 + (j-n)^2}{2\sigma^2}\right)$, where $\sigma$ represents the standard deviation of the Gaussian distribution. 

For the creation of SynDroneVision, square kernels with dimensions $M,N\in\{13,15,17,19,21\}$, $M=N$ are employed. The choice of kernel type and size is randomized independently for each image. Note that for kernel extensions beyond image boundaries  (i.e., when $x\notin\{0,...,W-1\}$ or $y\notin\{0,...,H-1\}$), explicit boundary conditions are imposed. Specifically, replication padding is employed to extend the image  boundaries by duplicating edge pixel values.

\subsection{Composition} 
For the generation of SynDroneVision, four to thirteen distinct camera positions are established per environment (cf. \Cref{tab:synthdata}). This environment-specific camera placement results in the capture of 72 annotated image sequences. Each sequence is characterized by a unique combination of drone model, lighting configurations, and background composition, providing a high degree of realism and variation. (An overview of selected camera field of views and annotation details are included in the supplementary material.)

SynDroneVision encompasses all 72 recorded image sequences, thus yielding a total of 131,307 annotated images for image-based drone detection (cf. \Cref{tab:synthdata}). Apart from drone images, the dataset also includes a small share of background images ($\sim$ 7\%). The dataset is partitioned into a training, validation, and test set. This allocation yields 119,307 images for training, 8,000 for validation, and 4,000 for testing purposes. The training and validation datasets are further augmented through the application of the aforementioned blurring technique (cf. \Cref{subsubsec:blur}). Specifically, $\sim$ 10\% of the images from each set are randomly selected, blurred, and included on top of the original data. This yields a final dataset size of 131,238 images for training and 8,800 images for validation  (cf. \Cref{tab:synthdata}). The number of test images remains unchanged.

\subsection{Characteristics}
\label{subsec:characteristics}
\begin{figure}[t!]
  \centering
  \includegraphics[width=0.475\textwidth, trim={0.2cm 22.5cm 0.9cm 0cm}, clip]{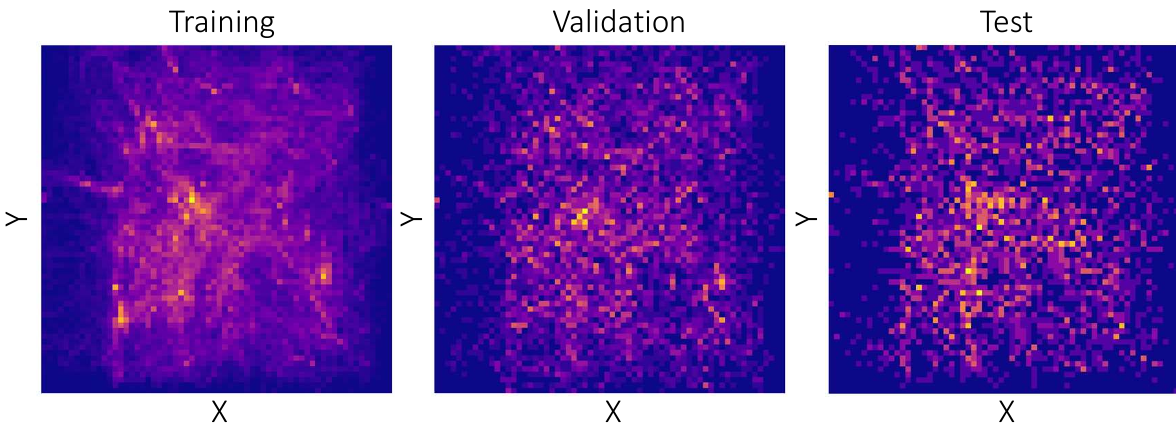}
  \caption{\label{fig:positions} Position distribution of drones within the SynDroneVision dataset. Regions of high frequency are shown in yellow, while areas with no data points are indicated in blue.}
\end{figure}

The proposed SynDroneVision dataset exhibits the following characteristic properties:

\begin{itemize}
    \item \textit{Image Resolution} -- An identical, consistently high resolution of 2560$\times$1489 pixels is maintained for all image sequences across all environments.
    \item \textit{Object Position} -- The spatial distribution of objects within the image frame, depicted in \Cref{fig:positions}, reveals a (mostly) uniform dispersion of objects across the entire image area, encompassing both central and peripheral image regions. This contrasts with the distribution patterns observed in datasets like DUT Anti-UAV~\cite{Jiang:2023} or MAV-VID~\cite{Rodriguez-Ramos:2020}, where objects are predominantly concentrated in the central region of the image.
\end{itemize}

\begin{table*}[btp]
\caption{Performance and technical details of the YOLOv8m, YOLOv8l, YOLOv9c, and YOLOv9e models evaluated on the DUT Anti-UAV dataset~\cite{Zhao:2022} (in-distribution data) across various training data configurations. The SynDroneVision dataset is abbreviated as SDV.}
  \label{tab:results}
  \centering
  \begin{tabular}{@{}|c|c|c|c|c|c|ccc|c|c|@{}}
    \hline
    YOLO &  Layers & Param. & GFLOPs & \multicolumn{2}{c|}{Training Data} & \multicolumn{5}{c|}{Evaluation on DUT Anti-UAV} \\
    & & {\small (M)} & {\small (B)} & {\small SDV  (Ours)} & {\small DUT Anti-UAV} & \multicolumn{3}{c|}{mAP~$\uparrow$} & FNR~$\downarrow$ & FDR~$\downarrow$\\
    &&&& {\footnotesize (synthetic)} & {\footnotesize  (real)} & {\small @0.25} & {\small @0.5} & {\small @0.5-0.95} & &\\\hline
    
    \multirow{3}{*}{v8m} & \multirow{3}{*}{295} & \multirow{3}{*}{25.85} & \multirow{3}{*}{\ \ 79.1} &  \ding{51} & -- & 0.677 & 0.639 & 0.422 & 0.525 & 0.103\\
    & & & & -- & \ding{51} & 0.956 & 0.933 & 0.669 & 0.118 & \textbf{0.021}\\
    & & & &  \ding{51} & \ding{51} & \textbf{0.960} & \textbf{0.938} & \textbf{0.686} & \textbf{0.117} &  0.025\\\hline
    
    \multirow{3}{*}{v8l} & \multirow{3}{*}{287} & \multirow{3}{*}{43.61} & \multirow{3}{*}{164.8} & 
    \ding{51} & -- & 0.746 & 0.716 & 0.468 & 0.438 & 0.079\\
    & & & & -- & \ding{51} & 0.922 & 0.896 & 0.628 & 0.149 & 0.067\\
    & & & & \ding{51} & \ding{51} & \textbf{0.963} & \textbf{0.944} & \textbf{0.696} & \textbf{0.105} & \textbf{0.015}\\\hline
    
    \multirow{3}{*}{v9c} & \multirow{3}{*}{384} & \multirow{3}{*}{25.32} & \multirow{3}{*}{102.3} &  \ding{51} & -- & 0.700 & 0.666&  0.429 & 0.474 & 0.093\\
    & & & &  -- & \ding{51} & 0.959 & 0.935 & 0.668 & 0.123 & 0.023\\
    & & & & \ding{51} & \ding{51} & \textbf{0.961} & \textbf{0.941} & \textbf{0.695} & \textbf{0.107} & \textbf{0.022} \\\hline
    
    \multirow{3}{*}{v9e} & \multirow{3}{*}{1225} & \multirow{3}{*}{58.15} & \multirow{3}{*}{192.7} & 
    \ding{51} & -- & 0.767 & 0.733 & 0.460 & 0.405 & 0.055\\
    & & & & -- & \ding{51} & 0.944 & 0.915 & 0.643 & 0.149 & 0.042 \\
    & & & & \ding{51} & \ding{51} & \textbf{0.969} & \textbf{0.955} & \textbf{0.723} & \textbf{0.095} & \textbf{0.009} \\\hline
  \end{tabular}
  
\end{table*}

\vspace*{-0.4cm}
\begin{itemize}
    \item \textit{Object Aspect Ratio} -- SynDroneVision exhibits considerable diversity in object aspect ratios. Minimum values range from 0.021 (train) to 0.041 (test), while maximum values range from 8.383 (test) to 9.993 (train). Similar to DUT Anti-UAV, most object aspect ratios fall between 1.0 and 3.0, with average values between 1.291 and 1.330 (cf. \Cref{tab:characteristics}). However, SynDroneVision covers a broader spectrum of aspect ratios compared to DUT Anti-UAV, which features aspect ratios ranging from 1 to 6.67 with an average of 1.91 (see \Cref{tab:characteristics}).
    
    \item \textit{Object Scale} -- The object scale, or object area ratio, quantifies the proportion of the drone area captured within the image frame relative to the total image area. Object scales within SynDroneVision exhibit a broad distribution, ranging from minimal values between 0.001~(train) and 0.009~(test) to maximum values of 1 (cf. \Cref{tab:characteristics}). However, the general trend leans towards smaller objects, with average values around 0.32 (cf. \Cref{fig:ratios}, supplementary material). This aspect is particularly crucial, given the prevalence of small drones in practical applications and the inherent complexity associated with their detection. It also aligns with the characteristic features observed in other drone detection datasets. For instance, the DUT Anti-UAV dataset also comprises predominantly small drone object, albeit with an average object scale of 0.013.
\end{itemize}

\section{Analysis}
\label{sec:experiment}
This section outlines the evaluation setup and presents key findings regarding the efficiency of SynDroneVision.

\subsection{Experimental Setup} 
To assess the effectiveness of SynDroneVision, we conduct a comparative analysis of various YOLO models and associated training configurations. Building upon the latest developments in YOLO architectures, our analysis focuses on YOLOv8~\cite{Varghese:2024} and YOLOv9~\cite{Wang:2024}, with a detailed examination of their respective variants: YOLOv8m, YOLOv8l, YOLOv9c, and YOLOv9e. Considering the  significance of real-world data in validating SynDroneVision's practical value, we incorporate the DUT Anti-UAV dataset~\cite{Zhao:2022} (selected for its characteristic resemblance to SynDroneVision). To ensure a comprehensive evaluation, especially in terms of model robustness, we also include the UAV-Eagle dataset~\cite{Barisic:2021} and the Drone Dataset by~\cite{Aksoy:2019} as out-of-distribution data.

The training procedure for each model involves three distinct strategies: (i)~training exclusively on SynDroneVision, (ii)~training solely on DUT Anti-UAV, and (iii)~a hybrid approach combining both datasets. Each model is trained for 100 epochs with a batch size of 64. For YOLOv9e, a reduced batch size of 32 is employed due to memory limitations. Other hyperparameters follow their default configurations specified in~\cite{Ultralytics}. Model performance is evaluated using the DUT Anti-UAV test set~\cite{Zhao:2022}, UAV-Eagle~\cite{Barisic:2021}, and the Drone Dataset by~\cite{Aksoy:2019}. Key performance indicators include standard object detection metrics such as mean average precision (mAP) at an intersection over union (IoU) threshold of 0.5, and computed over a range of IoU thresholds from 0.5 to 0.95. To account for precision variations in manually generated annotations (cf.~\cite{Lenhard:2024}), we also consider mAP values at an IoU threshold of 0.25, along with false negative rates (FNRs) and false discovery rates (FDRs). All experiments are performed using the Ultralytics repository~\cite{Ultralytics} on a single NVIDIA Quadro RTX-8000 GPU.

\begin{table*}[t!]
\caption{Performance of YOLOv9e on the UAV-Eagle dataset~\cite{Barisic:2021} and the Drone Dataset by~\cite{Aksoy:2019} (out-of-distribution data) across different training data configurations. The SynDroneVision dataset is abbreviated as SDV.}
  \label{tab:results2}
  \centering
  \begin{tabular}{@{}|l|c|c|ccc|c|c|@{}}
    \hline
    Evaluation Data& \multicolumn{2}{c|}{Training Data} &  \multicolumn{3}{c|}{mAP~$\uparrow$} & FNR~$\downarrow$ & FDR~$\downarrow$\\
     & {\small SDV  (Ours)} & {\small DUT Anti-UAV} & {\small @0.25} & {\small @0.5} & {\small @0.5-0.95} & &\\
    &{\footnotesize (synthetic)} & {\footnotesize  (real)} &&&&&\\\hline
    
    \multirow{2}{*}{UAV-Eagle~\cite{Barisic:2021}} & \ding{51} & -- & 0.946 & 0.810 & 0.340 & 0.193 & 0.125\\
    \multirow{2}{*}{{\footnotesize  (real)}} &  -- & \ding{51} & 0.947 & 0.819 & 0.290 & 0.165 & 0.088\\
    & \ding{51} & \ding{51} & \textbf{0.980} & \textbf{0.879} & \textbf{0.351} & \textbf{0.136} &  \textbf{0.058}\\\hline

    \multirow{2}{*}{Drone Dataset  by~\cite{Aksoy:2019}} & \ding{51} & -- & 0.741 & 0.474 & 0.179 & 0.379 & 0.109 \\
    \multirow{2}{*}{{\footnotesize  (real)}} &-- & \ding{51} & 0.812 & 0.569 & 0.202 & 0.251 & 0.185\\
    & \ding{51} & \ding{51} & \textbf{0.822} & \textbf{0.595} & \textbf{0.222} & \textbf{0.213} & \textbf{0.098} \\\hline
  \end{tabular}
  
\end{table*}

\subsection{Results} 
\label{subsec:results}
For evaluations on the SynDroneVision test set, refer to the supplementary material. In the following, we provide an assessment of all trained models on real-world data, highlighting SynDroneVision's practical applicability.

\vspace*{-0.2cm}
\paragraph*{Performance on DUT Anti-UAV.} 
A comprehensive analysis of YOLO models trained on the hybrid dataset combining SynDroneVision (synthetic data) and DUT Anti-UAV (real-world data) reveals substantial improvements across all performance indicators, particularly when compared to models trained solely on real-world data. Specifically, at an IoU threshold of 0.5, the hybrid training strategy yields an mAP increase of up to 4.8 percentage points over models trained exclusively on DUT Anti-UAV (cf. YOLOv8l, \Cref{tab:results}). This positive effect is even more pronounced for mAP values averaged across multiple IoU thresholds, resulting in improvements up to eight percentage points (cf. YOLOv9e, \Cref{tab:results}). Moreover, a considerable decline in FNRs and FDRs is observed across all models, except for YOLOv8m. The performance enhancements obtained by integrating SynDroneVision with real-world data are particularly amplified in more complex architectures characterized by an increased number of trainable parameters (e.g., cf. YOLOv8m vs. YOLOv8l, \Cref{tab:results}). The most effective performance is achieved with YOLOv9e.

A comparison between models trained exclusively on either SynDroneVision or DUT Anti-UAV also demonstrates distinct performance variations on the DUT Anti-UAV test set. Models trained on DUT Anti-UAV exhibit superior performance compared to those solely trained on SynDroneVision (cf. \Cref{tab:results}). In particular, models trained on SynDroneVision feature increased FNRs, ranging from 0.405 to 0.525 (cf. \Cref{tab:results}), while models trained on DUT Anti-UAV maintain FNRs below 0.149. This deviation is not surprising, considering the persistent challenge associated with the simulation-reality gap. Nevertheless, training exclusively on SynDroneVision still yields promising outcomes, with mAP values exceeding 0.639 at an IoU threshold of 0.5 for all YOLO architectures.

Visual examples of the detection results are provided in the supplementary material (\Cref{fig:resultsdetect1}).

\vspace*{-0.2cm}
\paragraph*{Performance on Out-of-Distribution Data.} The evaluation of YOLO models trained on SynDroneVision, both independently and in combination with real-world data, reveals notable robustness when exposed to varying data distributions and sources (cf. \Cref{tab:results2}). For instance, on the UAV-Eagle dataset, YOLOv9e trained on SynDroneVision achieves nearly equivalent performance to the model trained exclusively on DUT Anti-UAV, with mAP deviations of less than 0.01. It even exceeds the latter by five percentage points in mAP for an IoU threshold range of 0.5 to 0.95. Compared to corresponding validation outcomes on DUT Anti-UAV (see \Cref{tab:results}), the model trained solely on SynDroneVision shows significant improvements across all metrics, while the model trained exclusively on DUT Anti-UAV experiences notable performance declines (except for mAP at 0.25). Integrating SynDroneVision with DUT Anti-UAV during training further enhances performance, leading to even more pronounced improvements across key indicators. This trend is consistent across other YOLO variants, as detailed in the supplementary material (\Cref{tab:results3}). 

On the Drone Dataset by~\cite{Aksoy:2019}, the model solely trained on SynDroneVision exhibits lower performance in mAP and FNR relative to the DUT Anti-UAV-trained model. However, the performance gap is less pronounced than the one observed in \Cref{tab:results}. Combining both datasets during training also yields significant improvements on the Drone Dataset by~\cite{Aksoy:2019} across all metrics. An exception is the FDR, where the combination of SynDroneVision and DUT Anti-UAV is not always as effective for other YOLO variants (cf. \Cref{tab:results3}, supplementary material). 

\vspace*{-0.2cm}
\paragraph*{Discussion.} Our findings indicate that SynDroneVision is a highly promising dataset with the potential to substantially enhance the performance of DL models for drone detection (in surveillance applications), particularly when combined with real-world data. Furthermore, SynDroneVision seems to contribute to an improved model robustness, offering a clear benefit over exclusive real-world data training. The practical value of SynDroneVision is further highlighted by the promising performance of models trained exclusively on this dataset, considering that these models have not encountered target domain data or any real-world information. With its pixel-precise, automatically generated annotations, SynDroneVision significantly enhances bounding box localization, while simultaneously reducing data acquisition costs (without compromising performance).

\section{Conclusion}
\label{sec:conclusion}
In this work, we introduced SynDroneVision, a novel and comprehensive synthetic RGB dataset designed to support the development of robust drone detection systems. By providing a detailed analysis of recent YOLO models trained on both SynDroneVision and real-world data, we demonstrated the effectiveness of SynDroneVision in enhancing model accuracy and robustness. 

{\small
\bibliographystyle{ieee_fullname}
\bibliography{egbib}
}

\newpage
\twocolumn[{%
 \centering
 \Large \textbf{Supplementary Material} \\[3em]
}]

\appendix
\setcounter{table}{0}
\setcounter{figure}{0}
\renewcommand{\thetable}{\Roman{table}}
\renewcommand{\thefigure}{\Roman{figure}}
\section{Details on  Drone Detection Datasets}

\begin{table*}[t!]
\caption{Overview of additional characteristics of publicly available datasets for image-based drone detection. The symbols \ding{55} (does not apply) and \ding{51} (applies) indicate the designated computer vision (CV) task -- detection (detect) and~/~or tracking (track) -- the represented drone types (multicopter and~/~or fixed-wing), and the camera configurations (static and~/~or moving).}
  \label{tab:dronedatasetsdetails}
  \centering
  \begin{tabular}{@{}|l|c|c|p{2.8cm}|c|c|c|c|c|@{}}
    \hline
    Dataset & \multicolumn{2}{c|}{CV Task} & Objective &  \multicolumn{2}{c|}{Drone Type} &  \multicolumn{2}{c|}{Camera Config.}\\
    & {\small detect} &  {\small track} & & {\small multicopter} & {\small fixed-wing} & {\small static} & {\small moving}\\\hline
    
    USC  Drone Detect. \& Track.~\cite{Chen:2017,Wang:2018}  & \ding{51} & \ding{51} & drone monitoring & \ding{51} & \ding{55} &\ding{51} & \ding{51} \\\hline
    
    Drone Dataset~\cite{Aksoy:2019} & \ding{51} & \ding{55} & drone detection & \ding{51} & \ding{55} & -- &  --\\\hline
    
    MAV-VID~\cite{Rodriguez-Ramos:2020} & \ding{51} & \ding{51} & drone detection & \ding{51} & \ding{55} &  \ding{51} & \ding{51}\\\hline
    
    \multirow{2}{*}{Det-Fly~\cite{Zheng:2021}} & \multirow{2}{*}{\ding{51}} &  \multirow{2}{*}{\ding{55}} & detection of micro-UAVs & \multirow{2}{*}{\ding{51}} & \multirow{2}{*}{\ding{55}} &  \multirow{2}{*}{--} & \multirow{2}{*}{--}\\\hline
    
    \multirow{3}{*}{UAV-Eagle~\cite{Barisic:2021}} & \multirow{3}{*}{\ding{51}} & \multirow{3}{*}{\ding{51}} & UAV detection in unconstrained environments& \multirow{3}{*}{\ding{51}} & \multirow{3}{*}{\ding{55}} & \multirow{3}{*}{\ding{51}} & \multirow{3}{*}{\ding{55}} \\\hline
    
    UAVData~\cite{Zeng:2021} & \ding{51} & \ding{51} & UAV detection& \ding{51} & \ding{55} & \ding{51} & \ding{55}\\\hline
    
    \multirow{2}{*}{Halmstadt Data~\cite{Svanstroem:2021}} & \multirow{2}{*}{\ding{51}} & \multirow{2}{*}{\ding{51}}  & drone detection at airports & \multirow{2}{*}{\ding{51}} & \multirow{2}{*}{\ding{55}} & \multirow{2}{*}{\ding{51}} & \multirow{2}{*}{\ding{51}}\\\hline
    
    DUT Anti-UAV~\cite{Zhao:2022} & \ding{51} & \ding{51} & anti-UAV detection & \ding{51} & \ding{55} & \ding{51} & \ding{51}\\\hline
    
    \multirow{2}{*}{VisioDECT~\cite{Ajakwe:2022}} & \multirow{2}{*}{\ding{51}} & \multirow{2}{*}{\ding{55}} & detection of unauthorized UAVs & \multirow{2}{*}{\ding{51}} & \multirow{2}{*}{\ding{55}} & \multirow{2}{*}{\ding{51}} & \multirow{2}{*}{\ding{51}}\\\hline
    
    \multirow{2}{*}{Malicious Drones~\cite{Jamil:2022}} & \multirow{2}{*}{\ding{51}} & \multirow{2}{*}{\ding{55}} & hazardous payload drone detection & \multirow{2}{*}{\ding{51}} & \multirow{2}{*}{\ding{55}} & \multirow{2}{*}{--} & \multirow{2}{*}{--}\\\hline
    
    \multirow{2}{*}{S-UAV-T~\cite{Barisic:2022} \textit{(synthetic)}} & \multirow{2}{*}{\ding{51}} & \multirow{2}{*}{\ding{55}} & UAV-to-UAV detection& \multirow{2}{*}{\ding{51}} & \multirow{2}{*}{\ding{55}}& \multirow{2}{*}{\ding{55}} & \multirow{2}{*}{\ding{51}}\\\hline
    
    \multirow{2}{*}{Drone-vs-Bird Detection Ch.~\cite{Coluccia:2024}} & \multirow{2}{*}{\ding{51}} & \multirow{2}{*}{\ding{51}} & distinction between drones and birds & \multirow{2}{*}{\ding{51}} & \multirow{2}{*}{\ding{51}} &  \multirow{2}{*}{\ding{51}} & \multirow{2}{*}{\ding{51}}\\\hline
    
    \multirow{2}{*}{Anti-UAV~\cite{Jiang:2023}} & \multirow{2}{*}{\ding{55}} & \multirow{2}{*}{\ding{51}} & single UAV tracking & \multirow{2}{*}{\ding{51}} & \multirow{2}{*}{\ding{55}} & \multirow{2}{*}{\ding{51}} & \multirow{2}{*}{\ding{51}}\\\hline
  \end{tabular}
  
\end{table*}

In the field of image-based drone detection, diverse datasets have been established, each  defined by specific attributes and features tailored to distinct application contexts and objectives (see also \Cref{tab:dronedatasetsdetails}). Complementing the information in \Cref{subsec:datasets} (main paper), \Cref{tab:dronedatasetsdetails} and the subsequent sections offer a more comprehensive exploration of the  distinctive properties of each dataset.\vspace{-0.2cm}

\paragraph{USC Drone Detection and Tracking.} The USC Drone Detection and Tracking dataset~\cite{Chen:2017,Wang:2018} consists of 30 videos, each with a resolution of 1920$\times$1080 pixels. Recorded at a frame rate of 15 FPS and an approximate duration of one minute per video, the dataset contains approximately 27,000 images. The videos were captured on the University of Southern California (USC) campus, featuring a wide variety of backgrounds, camera angles, drone appearances, and diverse weather and lighting conditions. Only a single drone model (DJI Phantom) was used for data generation.\vspace{-0.2cm}

\paragraph*{Drone Dataset by ~\cite{Aksoy:2019}.} The Drone Dataset, provided by Aksoy \etal~\cite{Aksoy:2019}, comprises approximately 4,000 annotated RGB images sourced from YouTube drone videos and Google image searches. These images exhibit a resolution range from 300$\times$168 to 3840$\times$2160 pixels (4K) and exclusively feature DJI Phantom drones. The dataset also includes images of various non-drone objects.\vspace{-0.2cm}

\paragraph*{MAV-VID.} The Multirotor Aerial Vehicle VID (MAV-VID) dataset by Rodriguez-Ramos \etal~\cite{Rodriguez-Ramos:2020} comprises 64 videos (i.e., 40,232 images) of single drones. The videos are captured in various setups using different recording techniques, including handheld mobile devices, ground-based surveillance cameras, and other drones~\cite{IsaacMedina:2021}. The average drone size within the dataset is 136$\times$77 pixels.\vspace{-0.2cm}

\paragraph*{Det-Fly.} The Det-Fly dataset by Zheng \etal~\cite{Zheng:2021} focuses on air-to-air visual detection of micro UAVs and comprises 13,271 high-resolution images (3840×2160 pixels). The images, captured by another UAV, were sourced from videos at a sampling rate of 5 FPS or taken from selected positions. They feature diverse environmental backgrounds (sky, urban, field, mountain) and perspectives (front, top, bottom) based on relative viewing angles. Despite the considerable variability in drone size, with nearly half of the drones occupying less than 5\% of the total image area, the dataset  exclusively covers a single drone model (DJI Mavic).\vspace{-0.2cm}

\paragraph*{UAV-Eagle.} The UAV-Eagle dataset~\cite{Barisic:2021} is designed to evaluate the effectiveness of drone detection algorithms under varying conditions, including diverse illumination settings, motion artifacts, and viewpoint alterations. It comprises 510 annotated images featuring complex environments characterized by diverse background objects (e.g., trees, buildings, clouds, vehicles, and people). Employing a UAV-mounted camera for data collection, the dataset includes aerial images of both single- and multi-drone scenarios; however, limited to the Eagle quadcopter model.\vspace{-0.2cm}

\paragraph*{UAVData.} Zeng \etal~\cite{Zeng:2021} introduce UAVData, a dataset designed for visual drone detection, consisting of 13,803 manually recorded and annotated RGB images with a resolution of 1280$\times$720 pixels. The UAVData dataset captures a diverse array of real-world environments, encompassing both indoor settings (e.g., workshops and laboratories) and outdoor scenes featuring distinct background compositions (e.g., sky, trees, and buildings). This dataset aims to address the challenges inherent in real-world scenarios by incorporating rapid illumination changes, complex scenarios, and blurring effects caused by high-speed motion. In addition to six common drone models, UAVData includes balloon distractors, thus yielding 7,320 uni-drone images, 4,346 multi-drone images, and 2,137 balloon images. Drone sizes within the images range from 5$\times$23 to 720$\times$303 pixels. \vspace{-0.2cm}

\paragraph*{Halmstad Data.} The Halmstad Dataset, developed by Svanström \etal~\cite{Svanstroem:2021}, is a multi-sensor dataset for drone detection,  with a specific focus on detecting small UAVs. The dataset comprises 365 infrared (IR) and 285 visible light (RGB) videos, each lasting 10 seconds, alongside audio files. These recordings were primarily captured at airports in Sweden (e.g., Halmstad Airport) under daylight conditions. The dataset encompasses a variety of drone models (including the Hubsan H107D, DJI Flame Wheel, and DJI Phantom 4), as well as potential drone-like objects such as birds and airplanes. In total, the dataset comprises 203,328 annotated frames (across both IR and RGB), categorizing objects into the classes drone, bird, airplane, and helicopter. However, the .mat format annotations are not directly compatible with most DL frameworks.\vspace{-0.2cm}

\paragraph*{DUT Anti-UAV.} The Dalian University of Technology (DUT) Anti-UAV dataset~\cite{Zhao:2022} consists of two subsets: one for detection and one for tracking. The detection dataset includes 10,000 images, partially recorded in a sequential manner. The image resolutions vary significantly, ranging from 240$\times$160 to 5616$\times$3744 pixels. Object sizes within the images also exhibit substantial variation, with an average object area ratio of 0.013, indicating a high proportion of small objects. DUT Anti-UAV features 35 different UAV types for data generation and is characterized by a high diversity of scene information. It includes various outdoor environments such as the sky, dark clouds, jungles, high-rise buildings, residential buildings, farmland, and playgrounds. Additionally, it encompasses diverse lighting settings (day, night, dawn, and dusk) and weather conditions (sunny, cloudy, and snowy days). In terms of object positioning, the majority of drones are located in the central area of the image.\vspace{-0.2cm}

\paragraph*{VisioDECT.} The VisioDECT dataset~\cite{Ajakwe:2022} is a specialized aerial dataset designed for scenario-based detection of unauthorized drones. It comprises 20,924 annotated RGB images (852$\times$480 pixels) recorded across three distinct scenarios: cloudy, sunny, and evening. The images were captured at varying altitudes and locations, at different times, and under diverse climatic conditions, using six distinct drone models: Anafi Extended, DJI FPV, DJI Phantom, EFT-E410S, Mavic2-Air, and Mavic2-Enterprise Zoom. The collected data was manually cleaned (excluding images without drones) and annotated by domain experts.\vspace{-0.2cm}

\paragraph*{Malicious Drones.} Jamil \etal~\cite{Jamil:2022} introduce the Malicious Drones dataset, specifically designed for detecting harmful drones (e.g., carrying hazardous payloads) and differentiating them from other aerial entities. The dataset comprises 776 images categorized into five classes: aeroplane, bird, drone, helicopter, and malicious drone, with drones (normal and malicious) accounting for approximately half of the dataset ($\sim$ 399 images).  All images are standardized to a resolution of 224$\times$224 pixels. The dataset aims to address the complexity of real-world scenarios by including scenarios characterized by low illumination, reduced object visibility, occlusions, and adverse weather conditions.\vspace{-0.2cm}

\paragraph*{S-UAV-T.} The S-UAV-T dataset by Barisic et al.~\cite{Barisic:2022} is the only publicly available synthetic dataset for drone -- more precisely UAV-to-UAV -- detection. The dataset is generated via Blender~\cite{Blender} and the rendering engine Cycles~\cite{Cycles}, with a particular emphasis on texture randomization. To reflect the diversity of real-world environments, the dataset includes variations in drone models, the quantity of drones per image, lighting conditions (daylight, partly cloudy, twilight), object scales, camera positions and angles, as well as a range of unconventional textures. The dataset comprises 52,500 drone images with a resolution of 608$\times$608 pixels.\vspace{-0.2cm}

\paragraph*{Drone-vs-Bird Detection Challenge.} The Drone-vs-Bird Detection Challenge dataset~\cite{Coluccia:2024} is a comprehensive, manually annotated collection designed to assist in accurately distinguishing drones from birds across a wide range of conditions. It comprises 77 video sequences for training, each averaging 1,384 frames, captured with both static and moving cameras at resolutions from 720$\times$576 to 3840$\times$2160 pixels. The dataset includes eight types of commercial drones - three fixed-wing and five rotary-wing models - recorded in diverse environments such as urban areas, woodlands, agricultural fields, and maritime regions across Central Europe and the Mediterranean. These videos feature different weather conditions and times of day, introducing challenges like direct sun glare and varying camera characteristics. While drones are annotated, birds, which frequently appear as primary disturbance, are not. Drone sizes range from 15 pixels to over 1,000,000 pixels, with most annotated drones being smaller than 32$\times$32 pixels. The test set, comprises 30 additional video sequences without annotations, featuring new backgrounds, additional drone types, and other disturbing objects like planes.\vspace{-0.2cm}

\paragraph*{Anti-UAV.} The Anti-UAV dataset, created by Jiang \etal~\cite{Jiang:2023}, comprises 318 pairs of real RGB-T video sequences tailored for UAV tracking. Each pair features both RGB and thermal IR modalities, capturing a broad spectrum of lighting conditions (day and night) and diverse background compositions (e.g., buildings, clouds, or trees). Furthermore, the dataset includes prominent UAV models -- specifically the DJI Inspire, DJI Phantom 4, DJI Marvic Air, DJI Marvic Pro, DJI Spark, and Parrot. Similar to the DUT Anti-UAV dataset, the majority of drones in the Anti-UAV dataset are positioned centrally within the image frames. A comprehensive three-stage annotation process was used to generate precise annotations. The dataset does not specify a version explicitly dedicated to object detection.

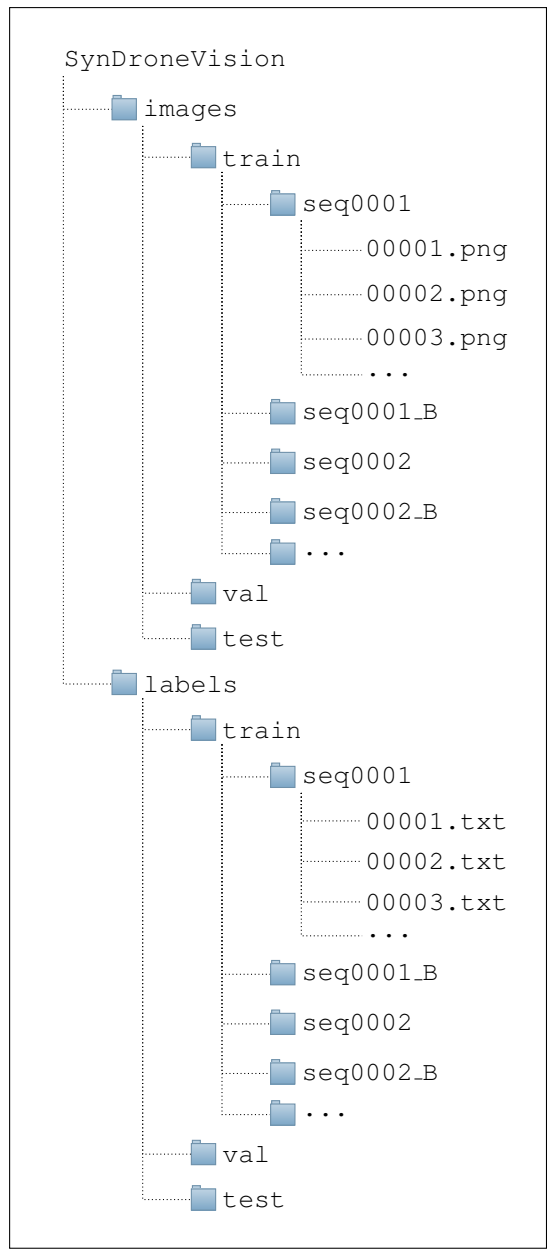
\begin{figure}[b!]
    \centering
\begin{tikzpicture}
\node[anchor=north west] (forest) at (0,0) {
\begin{forest}
    for tree={
                font=\ttfamily,
                grow'=0,
                folder indent=.7em, 
                folder icons,
                inner xsep=7pt,
                edge=densely dotted
            }
    [SynDroneVision   
                [images    
                    [train 
                        [seq0001 
                            [00001.png, is file] 
                            [00002.png, is file] 
                            [00003.png, is file]  
                            [..., is file]
                        ]
                        [seq0001\_B] 
                        [seq0002] 
                        [seq0002\_B] 
                        [...]
                    ]    
                    [val]    
                    [test]
                ]  
                [labels    
                    [train 
                        [seq0001 
                            [00001.txt, is file] 
                            [00002.txt, is file] 
                            [00003.txt, is file]  
                            [..., is file]
                        ]
                        [seq0001\_B] 
                        [seq0002] 
                        [seq0002\_B] 
                        [...]
                    ]       
                    [val]    
                    [test]
                ]
            ]
\end{forest}};
   
   \draw[draw] 
        ([shift={(-10pt, 10pt)}]forest.north west) 
        rectangle 
        ([shift={(10pt, -10pt)}]forest.south east);
\end{tikzpicture}

    \caption{Folder configuration of the SynDroneVision dataset.}
    \label{fig:datastructure}
\end{figure}

\section{Dataset Structure}
The dataset is structured into two main folders: \textit{images} and \textit{labels}. Each folders is further divided into training, test, and validation sets. Within these subdivisions, there are distinct folders for each image sequence, along with a subset of randomly blurred images (denoted by the suffix '\_B'). Annotations in the labels folder are provided as text files according to the YOLO standard format:~$$\texttt{<object-class> <x> <y> <width> <height>} \ \text{.}$$ Note  that $\texttt{<x>}$ and $\texttt{<y>}$ correspond to the normalized coordinates of the bounding box center. The normalization extends to both the  bounding box coordinates and dimensions. \Cref{fig:datastructure} shows SynDroneVision's structural organization.

\begin{table*}[t!]
\centering
\caption{Overview of environments used for synthetic data generation. The symbols \ding{51} (applies) and \ding{55} (does not apply) indicate an environment's public availability (2nd column from the left) and mandatory cost (3rd column from the left).}
  \label{tab:envs}
  \begin{tabular}{@{}|l|c|c|p{9cm}|@{}}
    \hline
    Environment & Publ. Avail. & Chargeable & Description\\\hline
    \multirow{4}{*}{University Site} & \multirow{4}{*}{\ding{55}} & \multirow{4}{*}{--} & A custom-designed environment replicating a German university campus situated within a wooded landscape. This urban setting features mid-rise building structures and a moderate vegetation density. \textbf{\Cref{fig:FOVs2}:} images 1--6 (rows 1--3). \\\hline
    
    \multirow{5}{*}{Venetian City~\cite{Unreal:Venice}} & \multirow{5}{*}{\ding{51}} & \multirow{5}{*}{\ding{51}} & A commercially available environment offering a realistic representation of Venice. The included demo map showcases Mediterranean-style buildings, autumnal trees, canals, stone bridges, and additional exterior elements such as benches and street lamps. \textbf{\Cref{fig:FOVs2}:} images 1--6 (rows 4--5). \\\hline
    
    \multirow{5}{*}{Farming Grounds~\cite{Unreal:Farming}} & \multirow{5}{*}{\ding{51}} & \multirow{5}{*}{\ding{51}} & A small agricultural environment featuring grain fields, fruit-bearing trees, and a variety of vegetable plants. In addition, the scene includes a small greenhouse, multiple raised garden beds, fencing, and other typical agricultural elements such as wooden barrels and crates. \textbf{\Cref{fig:FOVs2}:} images 1--6 (row 6).\\\hline
    
    \multirow{6}{*}{Rural Australia~\cite{Unreal:Australia}} & \multirow{6}{*}{\ding{51}} & \multirow{6}{*}{\ding{55}} & A publicly accessible environment capturing the expansive fields and open spaces characteristic of the Australian countryside. It includes detailed representations of natural elements, such as rivers, creeks, and rock formations, as well as native vegetation (e.g., shrubs and grasses) and local fauna (e.g., different bird species in flight). \textbf{\Cref{fig:FOVs2}:} images 1--6 (row 7).\\\hline
    
    \multirow{7}{*}{City Park~\cite{Unreal:CityPark}} & \multirow{7}{*}{\ding{51}} & \multirow{7}{*}{\ding{55}} & An urban park environment characterized by a rich diversity of lush vegetation, including trees, shrubs, flowers, and grass. The park features winding pathways and serene water features such as ponds and fountains. In addition to a few small buildings, the environment includes playgrounds, picnic areas, and sports grounds, as well as urban furniture such as benches, lampposts, and trash cans. \textbf{\Cref{fig:FOVs2}:} images 1--5 (row 8). \\\hline
    
    \multirow{8}{*}{Factory Grounds~\cite{Unreal:Factory}} & \multirow{8}{*}{\ding{51}} & \multirow{8}{*}{\ding{55}} &  An open-access environment showcasing a factory site. It exhibits various aspects of industrial architecture, including structures such as warehouses, production facilities, assembly lines, and storage installations, along with an extensive network of pipes, ducts, and other infrastructure. The environment also features a variety of machinery and equipment commonly found in factories or industrial settings. \textbf{\Cref{fig:FOVs2}:} image 6 (row 8), images 1--6 (row 9).\\\hline
    
    \multirow{7}{*}{Urban Downtown~\cite{Unreal:DowntownWest}} & \multirow{7}{*}{\ding{51}} & \multirow{7}{*}{\ding{55}} & A freely accessible environment featuring a Midwestern outdoor mall. Thus, the buildings are predominantly commercial, including shops, cafes, and restaurants. The urban design incorporates outdoor seating areas, green spaces, and playgrounds, set against a backdrop of mountains. The represented vegetation comprises flowers, small shrubs, and trees,  evoking a summer-like setting. \textbf{\Cref{fig:FOVs2}:} images 1--6 (row 10), images 1--4 (row 11).\\\hline
    
    \multirow{6}{*}{Modular Cityscape~\cite{Unreal:ModBuilding}} & \multirow{6}{*}{\ding{51}} & \multirow{6}{*}{\ding{55}} & An urban scene characterized predominantly by buildings (both commercial and residential) with diverse architectural styles. The environment integrates urban infrastructure, including streets and sidewalks, and is equipped with urban furniture such as benches, bus stops, streetlights, and trash receptacles. \textbf{\Cref{fig:FOVs2}:} images 5--6 (row 11), images 1--6 (last row)\\\hline
    \end{tabular}
    
\end{table*}

\begin{figure*}[ht!]
  \centering
  \includegraphics[width=\textwidth, trim={0.1cm 5.3cm 0.1cm 0cm}, clip]{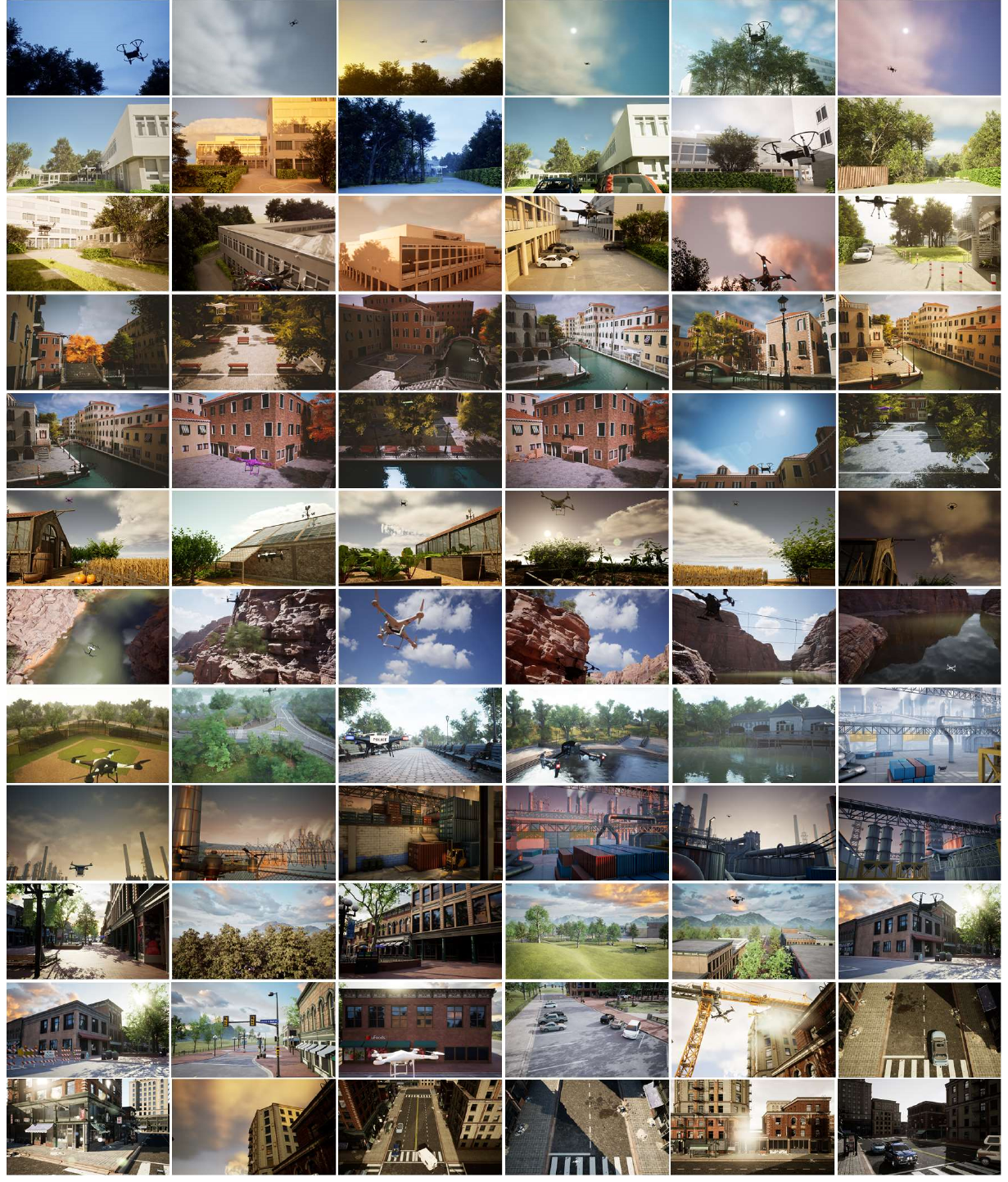}
  \caption{\label{fig:FOVs2} Customized camera perspectives and lighting configurations tailored to each environment. The camera fields of view (FOVs) correspond to the following environments (arranged from left to right, top to bottom): University Site (rows 1-3), Venetian City (rows 4-5), Farming Grounds (row 6), Rural Australia (row 7), City Park (images 1-5, row 8), Factory Grounds (image 6, row 8; images 1-6, row 9), Urban Downtown (images 1-6, row 10; images 1-4, row 11), and Modular Cityscape (images 5-6, row 11; images 1-6, last row).}
\end{figure*}

\begin{table*}[t!]
\caption{Parameter specifications for the Sun and Sky Actor to create environment-dependent illumination variations.}
  \label{tab:illumination_params1}
  \centering
  \begin{tabular}{@{}|l|c|c|>{\centering\arraybackslash}m{1.35cm}|>{\centering\arraybackslash}m{1.35cm}|c|c|c|c|c|c|@{}}
    \hline
    Environment & \multicolumn{2}{c|}{Solar Time} & \multicolumn{2}{c|}{Direct. Light Intensity} & \multicolumn{6}{c|}{Rayleigh Scattering (Channel Values)}\\
    & \multicolumn{2}{c|}{} & \multicolumn{2}{c|}{(lux)} & \multicolumn{2}{c|}{Red} & \multicolumn{2}{c|}{Green} & \multicolumn{2}{c|}{Blue}\\
    & {\small from} & {\small to} & {\small min} & {\small max} & {\small min} & {\small max} & {\small min} & {\small max} & {\small min} & {\small max} \\\hline
    
    University Site & 6.00 & 21.00 & 1 & 126 & 0.014 & 0.708 & 0.148 & 0.900 & 0.361 & 1 \\\hline
    Venetian City~\cite{Unreal:Venice}  & 9.40 & 17.00 & 6 & 15 & 0.100 & 0.565 & 0.199 &  0.739 & 0.410 & 0.990 \\\hline
    Farming Grounds~\cite{Unreal:Farming} & 7.24 & 14.25 & 5 & 13 & 0.119 & 0.599 & 0.188 & 0.836 & 0.361 & 1\\\hline
    Modular Cityscape~\cite{Unreal:ModBuilding} & 6.87 &  17.30 & 15 & 70 & 0.125 & 0.686 & 0.225 & 0.687 & 0.719 & 1 \\\hline
  \end{tabular}
  
\end{table*}

\section{Further Details on SynDroneVision}
\subsection{Environments}
The creation of SynDroneVision, as outlined in \Cref{subsec:datagen} (main paper), involved the application of diverse virtual environments. The majority of these environments are publicly available (some free of charge and some requiring payment). The only exception is the University Site environment, which was specifically designed to replicate a real-world scenario. \Cref{tab:envs} provides a comprehensive overview of the environments and their respective characteristics. \Cref{fig:FOVs2} illustrates camera perspectives and lighting configurations determined for each environment.

\subsection{Illumination Parameters}
To enhance the range of illumination within the SynDroneVision dataset, we primarily modified the settings of the Sun and Sky Actor~\cite{SunSkyActor} and the Post Process Volume~\cite{PostProcess}, essential tools within the Unreal Engine~\cite{Unreal}.
\vspace*{-0.2cm}
\paragraph*{Sun and Sky Actor.} For the Sun and Sky Actor, the following parameters were systematically modified:

\begin{itemize}
    \item \textit{Solar Time} -- The solar time parameter of the Sun and Sky Actor controls the position of the sun with respect to a pre-defined geographical location, simulating the natural progression of time during the day. Adjusting the solar time changes the sun's position relative to the horizon, creating different lighting conditions and shadows. 

    \item \textit{Directional Light Intensity} -- The intensity parameter of the Directional Light Actor controls the brightness of the light. Adjusting this parameter alters the overall illumination and shadow strength in the scene. Higher values increase brightness, while lower values decrease it.

    \item \textit{Rayleigh Scattering} -- The Rayleigh scattering parameter in Unreal's Sky Atmosphere contains both an RGB value and a scale. While the RGB value specifies the color tint of the scattering effect, the scale controls the overall intensity. This affects the sky's color and appearance, simulating natural atmospheric phenomena such as blue skies during the day and red hues at sunrise or sunset.
\end{itemize}
\noindent \Cref{tab:illumination_params1} summarizes the (environment-dependent) parameter value ranges employed in generating SynDroneVision.

\vspace*{-0.2cm}
\paragraph*{Post Process Volume.} To create variations in the scene's color grading, we refined the following color temperature-related parameters within the Post Process Volume: 
\begin{itemize}
    \item \textit{Temperature Type} -- The Temperature Type parameter specifies the method for adjusting the color temperature of a scene. Available options are White Balance (default) and Color Temperature. White Balance leverages the Temp value to calibrate the virtual camera, maintaining accurate white tones. Color Temperature utilizes the Temp value to directly adjust the scene's overall color hue. Both methods were employed in the generation process of SynDroneVision.

    \item \textit{Temp} -- The Temp parameter regulates the white balance relative to the scene’s light temperature. While higher values introduce a warm (yellow) coloration, lower values generate a cool (blue) tint. Matching temperature values ensure a neutral white light.
    
    \item \textit{Tint} -- The Tint parameter refines the white balance tint of a scene, correcting color imbalances to attain a more natural color representation across different light temperatures.
\end{itemize}
\noindent The parameter value ranges are detailed in \Cref{tab:illumination_params2}.

\begin{table}[t!]
  \caption{Post Process Volume settings.}
  \label{tab:illumination_params2}
  \centering
  \begin{tabular}{@{}|l|c|c|c|c|@{}}
    \hline
    Environment & \multicolumn{2}{c|}{Temp} & \multicolumn{2}{c|}{Tint}\\
    & {\small min} & {\small max} & {\small min} & {\small max}\\\hline
    University Site & 4,400 &  12,000 & 0 & 0.30 \\\hline
    Venetian City~\cite{Unreal:Venice}  & 3,840 & 15,000 & 0 & 0.25\\\hline
    Farming Grounds~\cite{Unreal:Farming} & 4,588 & 4,588 & 0.05 & 0.05\\\hline
    Modular Cityscape~\cite{Unreal:ModBuilding} & 4,770 & 9,500 & -0.02 & 0.03\\\hline
  \end{tabular}
\end{table}

\paragraph*{Rendering Settings.} The rendering settings of an Unreal project have a profound impact on both visual quality and system performance. In the generation process of SynDroneVision, we employed the rendering configurations specified in \Cref{tab:rendering_specs} for the majority of environments. Exceptions include the environments Factory Grounds~\cite{Unreal:Factory} and City Park~\cite{Unreal:CityPark}, which retained the default settings.

\begin{table}[t!]
\caption{Technical configuration details for Unreal projects.}
  \label{tab:rendering_specs}
  \centering
  \begin{tabular}{@{}|p{5.7cm}|c|@{}}
    \hline
    \multicolumn{2}{|l|}{\textbf{Global Illumination}} \\\hline
    Dynamic Global Illumination Methods & Lumen\\\hline
    
    \multicolumn{2}{|l|}{\textbf{Reflection}} \\\hline
    Reflection Method & Lumen\\
    Reflection Capture Resolution & 128\\
    Reduce Lightmap Mixing on Smooth Surfaces &  \multirow{2}{*}{\ding{51}}\\
    Support Global Clip Plane for Planar Reflections & \  \ \ \multirow{2}{*}{\ding{51}$^{\text{ \ \ding{72}}}$}\\\hline
    
    \multicolumn{2}{|l|}{\textbf{Lumen}} \\\hline
    Use Hardware Ray Tracing & \ding{51}\\
    Ray Lighting Mode & Surface Cache\\
    Software Ray Tracing Mode & Detail Tracing\\\hline
    
    \multicolumn{2}{|l|}{\textbf{Hardware Ray Tracing}} \\\hline
    Support Hardware Ray Tracing & \ding{51}\\
    Path Tracing & \ding{51}\\\hline
    
    \multicolumn{2}{|l|}{\textbf{Software Ray Tracing}} \\\hline
    Generate Mesh Distance Fields & \ding{51}\\
    Distance Field Voxel Density & 0.2\\\hline
    \multicolumn{2}{l}{{\footnotesize\ding{72}}{\small \ not enabled for University Site}
    }
  \end{tabular}
\end{table}

\subsection{Object Area Ratio and Object Aspect Ratio}
\begin{figure}[t!]
\centering
  \includegraphics[width=0.48\textwidth, trim={0.6cm 16.2cm 0.3cm 0cm}, clip]{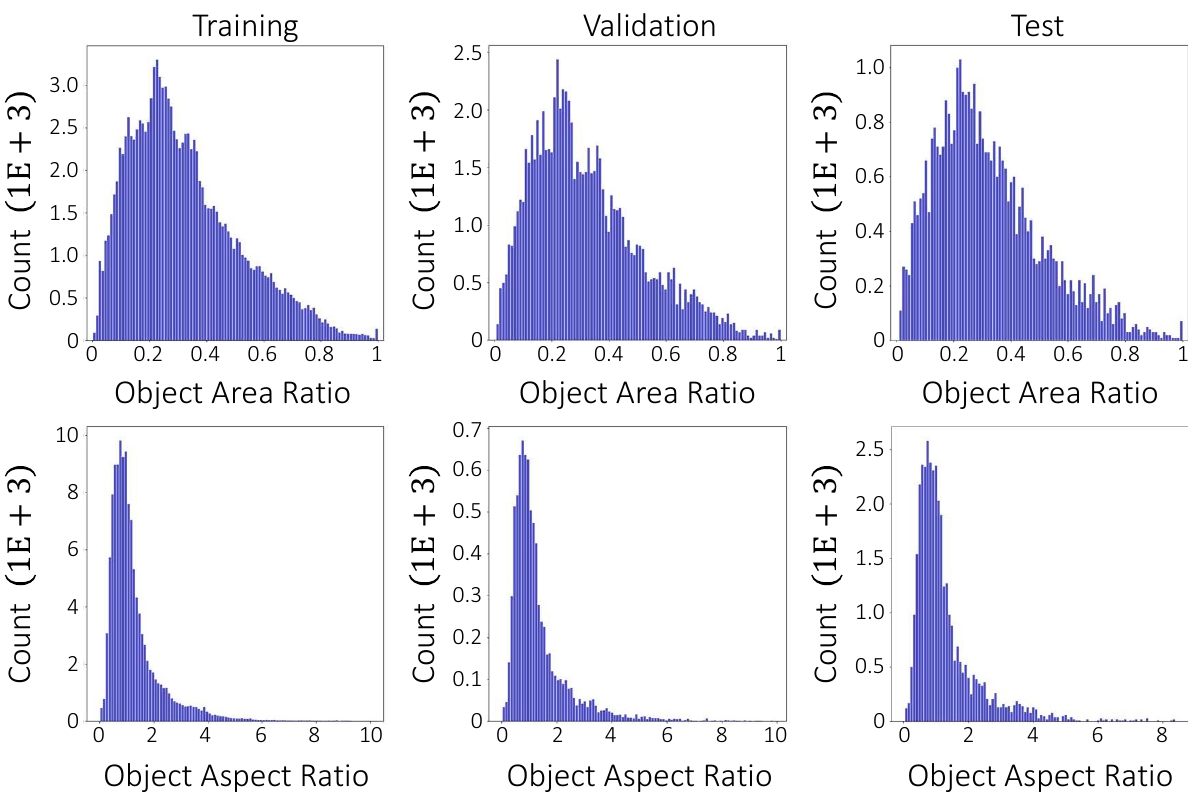}
  \caption{\label{fig:ratios} Object area and object aspect ratio distribution in the SynDroneVision dataset across training, validation, and test splits. }
\end{figure}

Supplementing the characteristics presented in \Cref{subsec:characteristics} (main paper), \Cref{fig:ratios} illustrates the distributions of object area (top) and object aspect ratios (bottom) for drones in the SynDroneVision dataset. Across all dataset partitions -- training, validation, and test -- the distribution of object area ratios exhibit a pronounced rightward skew. A comparable trend is observed in the distribution of aspect ratios.

\section{Analysis Details}
\subsection{Detection Examples}
\begin{figure}[t!]
\centering
 \includegraphics[width=0.49\textwidth, trim={0.2cm 18.2cm 8.9cm 0cm}, clip]{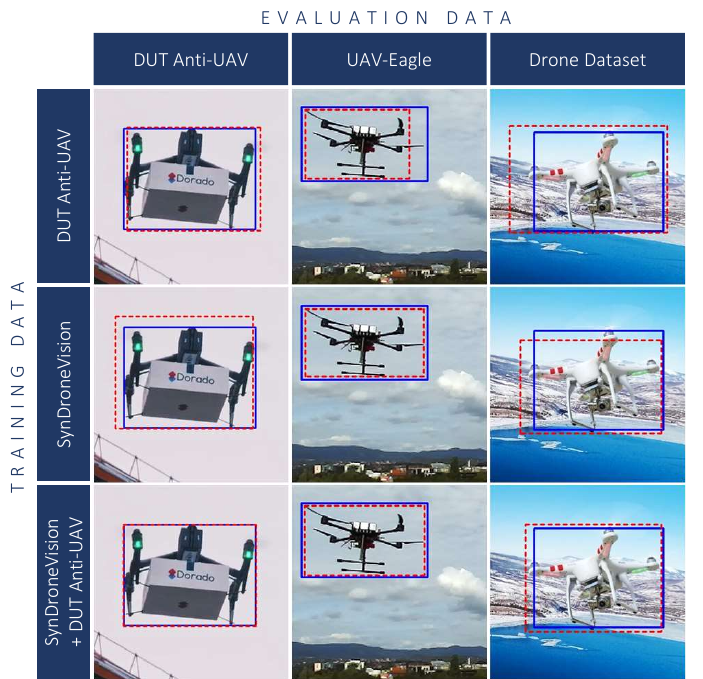}
 \caption{\label{fig:resultsdetect1} YOLOv9e detections on the DUT Anti-UAV test set~\cite{Zhao:2022} (1st column), the UAV-Eagle dataset~\cite{Barisic:2021}  (2nd column), and the Drone Dataset by~\cite{Aksoy:2019} (last column) demonstrating improved bounding box precision for models trained on both SynDroneVision and DUT Anti-UAV data (last row). Predictions (red dashed line) are marked alongside ground truth (solid blue line). }
\end{figure}

\begin{figure*}[tbp]
\centering
  \includegraphics[width=0.95\textwidth, trim={0.7cm 24.8cm 0.7cm 0cm}, clip]{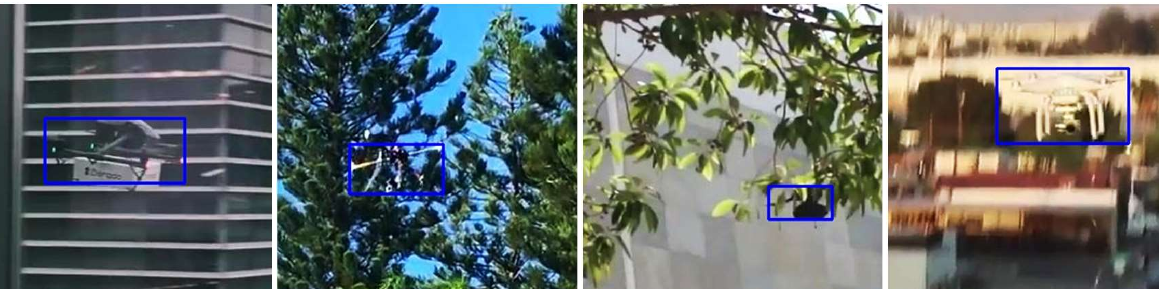}
  \caption{\label{fig:resultsdetect} Small cut-outs of selected samples from the DUT Anti-UAV test set that failed to be detected by YOLOv9e, irrespective of the employed training data. Ground truth bounding boxes are marked in blue.}
\end{figure*}

\Cref{fig:resultsdetect1} presents selected examples from the DUT Anti-UAV~\cite{Zhao:2022}, the UAV-Eagle~\cite{Barisic:2021}, and the Drone Dataset by~\cite{Aksoy:2019}, along with their corresponding detection outcomes obtained using YOLOv9e. The effectiveness of the detection results is compared across all three training strategies, i.e., YOLOv9e trained (i)~exclusively on SynDroneVision (first row), (ii)~solely on DUT Anti-UAV  (second row), and (iii)~on a combination of  both datasets (last row). The figure illustrates the superior bounding box localization achieved by the strategic combination of both datasets during training, supporting the significant performance enhancements in mAP values discussed in \Cref{subsec:results} (main paper). Conversely, \Cref{fig:resultsdetect} displays selected examples from DUT Anti-UAV where YOLOv9e fails to detect existing drones, irrespective of the training data.

\begin{table*}[tbp]
\caption{Performance of YOLOv8m, YOLOv8l, and YOLOv9c on the UAV-Eagle dataset~\cite{Barisic:2021} and the Drone Dataset by~\cite{Aksoy:2019} (out-of-distribution data) across different training data configurations. The SynDroneVision dataset is abbreviated as SDV.}
  \label{tab:results3}
  \centering
  \begin{tabular}{@{}|l|c|c|c|ccc|c|c|@{}}
    \hline
    Evaluation Data & YOLO & \multicolumn{2}{c|}{Training Data} &  \multicolumn{3}{c|}{mAP~$\uparrow$} & FNR~$\downarrow$ & FDR~$\downarrow$\\
    & & {\small SDV  (Ours)} & {\small DUT Anti-UAV} & {\small @0.25} & {\small @0.5} & {\small @0.5-0.95} & &\\
    & &{\footnotesize (synthetic)} & {\footnotesize  (real)} &&&&&\\\hline
    
    \multirow{8}{*}{UAV-Eagle~\cite{Barisic:2021}} & \multirow{3}{*}{v8m} & \ding{51} & -- & 0.944 & 0.771 & 0.293 & 0.201 & 0.169 \\
    \multirow{8}{*}{{\footnotesize  (real)}} & & -- & \ding{51} & 0.935 &  0.823 & 0.302 & 0.199 & \textbf{0.063}\\
    & & \ding{51} & \ding{51} & \textbf{0.961} & \textbf{0.849} & \textbf{0.350} & \textbf{0.136} &  0.089\\\cline{2-9}
    
    & \multirow{3}{*}{v8l} & \ding{51} & -- & 0.951 & 0.786 & 0.304 & 0.217 & 0.125 \\
    & & -- & \ding{51} & 0.920 & 0.725 & 0.217 & 0.180 & 0.224\\
    & & \ding{51} & \ding{51} & \textbf{0.979} & \textbf{0.869} & \textbf{0.368} & \textbf{0.126} & \textbf{0.074} \\\cline{2-9}
    
    & \multirow{3}{*}{v9c} & \ding{51} & -- & 0.926 & 0.770 & 0.289  & 0.216 & 0.163\\
    & & -- & \ding{51} & 0.922 & 0.799 & 0.275 & 0.219 & \textbf{0.077}\\
    & & \ding{51} & \ding{51} & \textbf{0.975} & \textbf{0.859} & \textbf{0.353} & \textbf{0.141} & 0.092\\\hline
    \multirow{8}{*}{Drone Dataset by~\cite{Aksoy:2019}} & \multirow{3}{*}{v8m} & \ding{51} & -- & 0.758 & 0.527 & 0.188 & 0.310 & 0.138\\
    \multirow{8}{*}{{\footnotesize  (real)}} & & -- & \ding{51} & 0.801 & 0.560 & 0.208 & 0.278 & \textbf{0.113}\\
    & & \ding{51} & \ding{51} & \textbf{0.824} & \textbf{0.613} & \textbf{0.232} & \textbf{0.196} & 0.114 \\\cline{2-9}
    
    & \multirow{3}{*}{v8l} & \ding{51} & -- & 0.768 & 0.515 & 0.193 & 0.389 & \textbf{0.076}\\
    & & -- & \ding{51} & \textbf{0.800} & 0.552 & 0.199 & 0.263 & 0.227\\
    & & \ding{51} & \ding{51} & 0.799 & \textbf{0.603} & \textbf{0.227} & \textbf{0.216} & 0.116 \\\cline{2-9}
    
    & \multirow{3}{*}{v9c} & \ding{51} & -- & 0.737 & 0.530 & 0.199 & 0.401 & \textbf{0.073}\\
    && -- & \ding{51} & 0.806 & 0.556 & 0.206 & 0.292 & 0.134\\
    & & \ding{51} & \ding{51} & \textbf{0.825} & \textbf{0.606} & \textbf{0.224} & \textbf{0.198} & 0.126  \\\hline
  \end{tabular}
    
\end{table*}

\subsection{Performance on Out-of-Distribution Data} 
Section 4.2 (main paper) highlights that the performance and robustness enhancements achieved with SynDroneVision on out-of-distribution data are not limited to YOLOv9e, but extend to other YOLO variants as well. Evaluating YOLOv8m, YOLOv8l, and YOLOv9c on the UAV-Eagle dataset also demonstrates that training exclusively with either SynDroneVision or DUT Anti-UAV yields comparably strong results across all performance indicators (see \Cref{tab:results3}). In some cases, models trained solely on SynDroneVision perform even better than those trained on real-world data, particularly in terms of mAP values at an IoU threshold of 0.25. In analogy to YOLOv9e, the best performance is achieved when combining both datasets during training. Here, YOLOv8l exhibits the most significant improvement over exclusive real-world data training, featuring a 14.4 percentage point increase in mAP at an IoU threshold of 0.5 and a 10.51 percentage point improvement across a range of IoU thresholds from 0.5 to 0.95 (cf. \Cref{tab:results3}).  Furthermore, integrating synthetic and real-world data effectively lowers the FNR, whereas variations in the FDR remain inconsistent.

For the Drone Dataset by~\cite{Aksoy:2019}, models trained exclusively on SynDroneVision exhibit slightly lower mAP values compared to those trained solely on DUT Anti-UAV. Nevertheless, the integration of both datasets yields overall performance enhancements, as detailed in \Cref{tab:results3}. The only exception seems to be YOLOv8l, where the mAP value at an IoU threshold of 0.25 is marginally higher for the model trained exclusively on DUT Anti-UAV. However, the discrepancy is negligible, with a difference of only 0.001.

\subsection{Performance on SynDroneVision} 
To provide a comprehensive understanding of model performance, we also incorporate the SynDroneVision test set into our evaluation. Specifically, we focus on models trained either exclusively on SynDroneVision or on a combination of SynDroneVision and DUT Anti-UAV. \Cref{tab:resultsSynth} highlights the consistently high performance of the models across all performance indicators.

\begin{table*}[t!]
\caption{Performance of YOLOv8m, YOLOv8l, YOLOv9c, and YOLO9e on the SynDroneVision test set across different training data configurations. The SynDroneVision dataset is abbreviated as SDV.}
  \label{tab:resultsSynth}
  \centering
  \begin{tabular}{@{}|c|c|c|ccc|c|c|@{}}
    \hline
    YOLO & \multicolumn{2}{c|}{Training Data} &  \multicolumn{5}{c|}{Evaluation on SynDroneVision} \\
    & {\small SDV  (Ours)} & {\small DUT Anti-UAV} & \multicolumn{3}{c|}{mAP~$\uparrow$} & FNR~$\downarrow$ & FDR~$\downarrow$\\
    &{\footnotesize (synthetic)} & {\footnotesize  (real)} & {\small @0.25} & {\small @0.5} & {\small @0.5-0.95} &&\\\hline
    
    \multirow{2}{*}{v8m} & \ding{51} & -- & 0.995 & 0.995 & 0.944 & 0.013 & 0\\
    & \ding{51} & \ding{51} & 0.995 & 0.995 & 0.942 & 0.014 & 0\\\hline
    
    \multirow{2}{*}{v8l} & \ding{51} & -- & 0.995 & 0.995 & 0.955 & 0.014 & 0.001\\
    & \ding{51} & \ding{51} & 0.995 & 0.995 & 0.956 & 0.013 & 0\\\hline
    
    \multirow{2}{*}{v9c} & \ding{51} & -- & 0.995 & 0.995 & 0.952 & 0.014 & 0.001\\
    & \ding{51} & \ding{51} & 0.995 & 0.995 & 0.954 & 0.014 & 0\\\hline
    
    \multirow{2}{*}{v9e} & \ding{51} & -- & 0.995 & 0.995 & 0.967 & 0.014 & 0.001\\
    & \ding{51} & \ding{51} & 0.995 & 0.995 & 0.967 & 0.014 & 0\\\hline
  \end{tabular}
\end{table*}
 
\vspace*{20cm}

\end{document}